\documentclass[a4paper,fleqn]{cas-dc}

\usepackage[numbers]{natbib}
\usepackage{caption}
\usepackage{makecell}
\usepackage{subcaption}
\usepackage{tabularx}
\usepackage{longtable}

\def\tsc#1{\csdef{#1}{\textsc{\lowercase{#1}}\xspace}}
\tsc{WGM}
\tsc{QE}
\begin{document}
\let\WriteBookmarks\relax
\def\floatpagepagefraction{1}
\def\textpagefraction{.001}

\shorttitle{Context-Aware Attention Layers coupled with Optimal Transport Domain Adaptation and Multimodal Fusion methods for recognizing dementia from spontaneous speech}

\shortauthors{L.Ilias, D.Askounis}  

\title [mode = title]{Context-Aware Attention Layers coupled with Optimal Transport Domain Adaptation and Multimodal Fusion methods for recognizing dementia from spontaneous speech}  

%\tnotemark[<tnote number>] 

%\tnotetext[<tnote number>]{<tnote text>} 


\author[1]{Loukas Ilias}[orcid=0000-0002-4483-4264]

\cormark[1]

%\fnmark[<footnote mark no>]

\ead{lilias@epu.ntua.gr}

%\ead[url]{<URL>}

\credit{Conceptualization, Methodology, Software, Validation, Formal Analysis, Visualization, Investigation, Data Curation, Writing - Original Draft, Writing - Review \& Editing}

\affiliation[1]{organization={Decision Support Systems Laboratory, School of Electrical and Computer Engineering, National Technical University of Athens},
            addressline={Zografou}, 
            city={Athens},
%          citysep={}, % Uncomment if no comma needed between city and postcode
            postcode={15780}, 
            %state={},
            country={Greece}}

\author[1]{Dimitris Askounis}[orcid=0000-0002-2618-5715
]

% Footnote of the second author

% Email id of the second author
\ead{askous@epu.ntua.gr}


\credit{Supervision, Project Administration, Writing - Review \& Editing}


%\affiliation[<aff no>]{organization={},
 %           addressline={}, 
 %           city={},
%          citysep={}, % Uncomment if no comma needed %between city and postcode

\cortext[1]{Corresponding author}


\begin{abstract}
Alzheimer's disease (AD) constitutes a complex neurocognitive disease and is the main cause of dementia.  Although many studies have been proposed targeting at diagnosing dementia through spontaneous speech, there are still limitations. Existing state-of-the-art approaches, which propose multimodal methods, train separately language and acoustic models, employ majority-vote approaches, and concatenate the representations of the different modalities either at the input level, i.e., early fusion, or during training. Also, some of them employ self-attention layers, which calculate the dependencies between representations without considering the contextual information. In addition, no prior work has taken into consideration the model calibration. To address these limitations, we propose some new methods for detecting AD patients, which capture the intra- and cross-modal interactions. First, we convert the audio files into log-Mel spectrograms, their delta, and delta-delta and create in this way an image per audio file consisting of three channels. Next, we pass each transcript and image through BERT and DeiT models respectively. After that, context-based self-attention layers, self-attention layers with a gate model, and optimal transport domain adaptation methods are employed for capturing the intra- and inter-modal interactions. Finally, we exploit two methods for fusing the self and cross-attention features. For taking into account the model calibration, we apply label smoothing. We use both performance and calibration metrics. Experiments conducted on the ADReSS and ADReSSo Challenge datasets indicate the efficacy of our introduced approaches over existing research initiatives with our best performing model reaching Accuracy and F1-score up to 91.25\% and 91.06\% respectively. 
\end{abstract}


%\item 
%\item 
%\item 
%end{highlights}

% Keywords

\maketitle

\section{Introduction}

Alzheimer's disease (AD) is a progressive neurologic disorder and constitutes the most common cause of dementia. According to the World Health Organization, around 55 million people have dementia worldwide with over 60\% living in low- and middle-income countries \cite{alzheimer_who}. In addition, dementia affects the ability of a person to communicate. More specifically, people with dementia may not be capable of finding the right words or may not be able to find any word at all. Concurrently, they are not able to stay focused on a discussion and tend to use words without meaning, thus being unable to communicate with other people \cite{dementia_language}. This fact entails physical, psychological, social, and economic impacts not only  for people living with dementia, but also for their carers, families, and society at large. Due to the fact that dementia becomes worse over time, it is important to be diagnosed early. For this reason, there have been several studies proposed, which distinguish AD patients from non-AD ones using speech and transcripts. 

Although many methods have been proposed for fusing representation vectors from different modalities in many tasks, the task of multimodal dementia detection using speech and transcripts has still substantial limitations. More specifically, most previous research works exploit early fusion, late fusion approaches, or add/concatenate the representations obtained by the different modalities. For the early fusion approaches, the authors simply concatenate features from the different modalities at the input level \cite{pompili20_interspeech, martinc20_interspeech}. For the late fusion approaches, the authors train textual and acoustic models separately and perform final decision voting in a weighted manner \cite{mittal2021multimodal}. Other approaches train textual and acoustic models separately and apply a majority-vote approach for the final prediction \cite{cummins2020comparison}. It is obvious that these approaches do not account for both intra- and inter-modality interactions. Concurrently, majority-vote approaches increase the training time, since multiple models must be trained and tested separately. In addition, early fusion strategies or the add/concatenation operation give equal importance to the different modalities and do not capture the inherent correlations between the two modalities.

Recently, some studies \cite{10.3389/fnagi.2022.830943,9926818} address the limitations of fusing the different modalities. However, some limitations still exist. Specifically, the work in \cite{10.3389/fnagi.2022.830943} concatenates the representation vectors of the two modalities and exploits a self-attention layer incorporating a gated model. However, in terms of the textual modality recent studies have shown that Self-Attention layers treat the input sequence as a bag-of-word tokens and each token individually performs attention over the bag-of-word tokens. Consequently, the contextual information is not taken into account in the calculation of dependencies between elements. There have been proposed a number of studies enhancing the self-attention layers with contextual information \cite{tu-etal-2017-context,8031316,wang-etal-2017-exploiting-cross,voita-etal-2018-context}. In addition, the study in \cite{9926818} employs a tensor fusion layer. However, the output tensor is high dimensional.

In addition, the reliability of a machine learning model’s confidence in its predictions, denoted as calibration \cite{doi:10.1080/01621459.1982.10477856,VerificationofProbabilisticPredictionsABriefReview}, is critical for high risk applications, such as deciding whether to trust a medical diagnosis prediction \cite{doi:10.1177/0962280213497434,10.1136/amiajnl-2011-000291,pmlr-v97-raghu19a}. However, no prior work has taken into account the calibration of the models, creating in this way overconfident models. According to \cite{pmlr-v70-guo17a}, modern neural networks are not well-calibrated, while they are overconfident at the same time.

In order to tackle the aforementioned limitations, we introduce deep neural networks, which are trained in an end-to-end trainable manner and capture both the inter- and intra-modal interactions. First, we convert the audio files into images consisting of three channels, namely log-Mel spectrograms, their delta, and delta-delta. Next, each transcript and image are passed through BERT \cite{devlin-etal-2019-bert} and DeiT \cite{pmlr-v139-touvron21a} models respectively. In order to ensure that the sequence length of the vectors obtained by BERT and DeiT is the same, we exploit an Optimal Transport Kernel (OTK) Embedding. We pass the textual representation through an enhanced self-attention layer with contextual information. We exploit three main methods for the contextualization, including the global context, deep context, and deep-global context \cite{CHEN2022108980,Yang_Li_Wong_Chao_Wang_Tu_2019}. Next, we pass the image representation through a self-attention mechanism with a novel gating model proposed by \cite{yu2019multimodal} to model the intra-modal interactions. Motivated by the study of \cite{Pramanick_2022_WACV}, we use optimal transport based domain adaptation \cite{villani2008optimal} methods for capturing the inter-modal interactions. Then, we propose two attention-based methods for fusing the self and cross-attention features. Finally, for preventing models becoming too overconfident, we use label smoothing. We use metrics for assessing both the performance and the calibration of our model. We verify the effectiveness and robustness of our approaches by conducting experiments on two publicly available datasets, namely ADReSS and ADReSSo Challenge datasets, and using both manual and automatically generated transcripts. We show that our introduced approaches obtain multiple advantages over the state-of-the-art approaches.

Our main contributions can be summarized as follows:
\begin{itemize}
    \item To the best of our knowledge, this is the first study utilizing DeiT, optimal transport kernel, and optimal transport domain adaptation methods in the task of dementia detection from spontaneous speech.
    \item This is the first study in the task of dementia detection from spontaneous speech exploiting label smoothing for preventing the models become too overconfident. We also evaluate our proposed models in terms of both the performance and the calibration.
    \item This is the first study in the task of dementia detection from speech data exploiting context-aware self-attention mechanisms and comparing two different approaches for fusing the self- and cross-attention features.
    \item We conduct a series of ablation experiments to demonstrate the effectiveness of the introduced approach. We evaluate our approaches on the ADReSS and ADReSSo Challenge datasets and show that they achieve competitive results to the existing research initiatives.
\end{itemize}

\section{Related Work}

\subsection{Dementia Detection}

\subsubsection{Unimodal Approaches}

Bertini et al. \cite{BERTINI2022101298} introduced a unimodal approach using only speech data to detect AD patients. First, the authors applied a data augmentation technique, namely \textit{SpecAugment} \cite{park19e_interspeech}, for increasing the size of the data. Next, the authors used an autoencoder, called \textit{auDeep} \cite{freitag2017audeep}, and passed the latent vector representation through a multilayer perceptron for the final prediction. The main limitation of this approach is pertinent to the representation of the speech signal as a log-Mel spectrogram. However, appending delta and delta-delta features to log-Mel spectrogram is proven to be more beneficial, since delta features add dynamic information \cite{5947425}.

The research work proposed by \cite{10.3389/fpsyg.2020.624137} employed unimodal approaches by using only either speech or text to classify subjects into AD patients or non-AD ones. For the text modality, the authors extracted embeddings by using fastText, BERT, LIWC, and CLAN. For the acoustic modality, the authors extracted i-vectors and x-vectors. For both modalities, they employed dimensionality reduction techniques and trained shallow machine learning classifiers and neural networks (CNNs and LSTMs). The authors claimed that the Support Vector Machine and the Random Forest Classifiers trained on BERT embeddings achieved the highest accuracy. One limitation of this study is the fact that the authors used BERT embeddings as features for training additional algorithms. They did not experiment with extracting the [CLS] token and passing it to a dense layer for performing the classification.

Karlekar et al. \cite{karlekar-etal-2018-detecting} applied three deep neural networks based on CNNs, LSTM-RNNs, and their conjunction to distinguish AD patients from non-AD ones utilizing only transcripts. Next, they proposed explainability techniques by applying automatic cluster pattern analysis and first derivative saliency heat maps, in order to uncover differences in language between AD patients and healthy control groups. The main limitation of this paper is the fact that the authors did not experiment with language models based on transformers, i.e., BERT, RoBERTa, and so on.

The authors in \cite{10.1145/3175587.3175589} extracted a large number of acoustic features for detecting people in healthy control, mild cognitive impairement, and AD groups and used a longitudinal dataset for conducting their experiments. After extracting the acoustic features, they applied feature selection techniques for finding the optimal set of features. Also, they applied sampling techniques, namely the Synthetic Minority
Over-sampling Technique, for dealing with the imbalanced dataset. The Support Vector Machine and the Stochastic Gradient Descent were used for the classification purposes. The limitation of this paper lies on the feature extraction process, which is a time-consuming and tedious procedure. Additionally, the optimal feature set may not be found, since some level of domain expertise is required.

Similarly, the work proposed by \cite{6864431} extracted seventeen features from transcripts for detecting AD patients. Specifically, the authors extracted the rate of pauses in utterances, filler sounds, number of no answers, part-of-speech tags, intelligibility of speech, diversity and complexity of the words, and many more. Next, they trained Support Vector Machines, Linear Discriminant Analysis, and Decision Trees. Results indicated that 90\% prediction accuracy can be obtained using only phone entropy, silence rate per utterance, and word entropy with a Decision Tree classifier. The limitation of this paper lies on the feature extraction process, which is a time-consuming and tedious procedure. Additionally, the optimal feature set may not be found, since some level of domain expertise is required.

An augmented adversarial self - supervised learning method was proposed by \cite{yang22k_interspeech}. Specifically, the introduced approach was based on contrastive predictive encoding. For dealing with the imbalanced dataset, i.e., limited number of speech samples corresponding to AD patients, the authors applied three augmentation schemes, including speed based augmentation, tempo based augmentation, and tremolo based augmentation. Findings indicated that the proposed methods improved the performance for AD detection to a large margin compared to other models.

\subsubsection{Multimodal Approaches}

Several approaches have been introduced which fuse the representation vectors or features of the different modalities at the input level. This strategy is known as an early fusion approach and does not capture effectively the inter-modal interactions. Edwards et al. \cite{edwards20_interspeech} proposed a multimodal (audio and text) and multiscale (word and phoneme levels) approach. For the acoustic modality, the authors extracted features using the OpenSMILE toolkit, applied feature selection techniques, and trained shallow machine learning classifiers, including SVM, latent discriminant analysis (LDA), and LR. In terms of the language models, the authors trained a Random Forest Classifier on Word2Vec and GloVe embeddings. Also, they trained a FastText classifier from scratch. In addition, pretrained embeddings obtained by Sent2Vec, RoBERTa, ELECTRA, and so on were fine-tuned with the FastText classifier. The authors transcribed the segmented text into phoneme written pronunciation
using CMUDict and stated that the FastText classifier was the best performing model trained on the phoneme representation. Results also showed that the combination of phonemes and audio yielded to the highest accuracy accounting for 79.17\%. Martinc and Pollak \cite{martinc20_interspeech} proposed also an early fusion approach. The authors extracted a large number of features corresponding to the textual and acoustic modality. They fused the feature sets via an early fusion method. Finally, they trained four machine learning classifiers, namely XGBoost, Random Forest, SVM, and Logistic Regression. Findings showed that the logistic regression and SVMs were proved to be better than XGBoost and Random Forest. Also, the authors stated that the readability features led to a surge in the classification performance. In terms of the audio features, the duration was the best performing one. Pompili et al. \cite{pompili20_interspeech} proposed an early fusion approach for fusing the modalities of speech and transcript. Specifically, for the text modality, the authors employed the BERT model first and then trained three deep neural models on top of the BERT embeddings, namely (i) a Global Maximum pooling, (ii) a bidirectional LSTM-RNNs provided with an attention module, and (iii) the second model augmented with part-of-speech (POS) embeddings. For the audio modality, the authors extracted the x-vectors. Finally, the authors merged the feature sets corresponding to the two different modalities and trained a Support Vector Machine classifier. Results showed that the fusion of the two modalities increased the performance obtained by unimodal approaches exploiting only speech or text.

Other approaches employ late-fusion strategies. This means that multiple models, i.e., acoustic and language, are trained separately and the final result/prediction is often taken after a majority vote approach. In this way, the inter-modal interactions are not captured. The authors in \cite{9459113} proposed a majority-level approach for classifying AD patients using the audio and textual modalities. In terms of the textual modality, the authors extracted handcrafted textual features and deep textual embeddings of transcripts. For the extraction of deep textual embeddings, they used BERT, RoBERTa, and distilled versions of BERT and RoBERTa. Next, they exploited feature aggregation techniques and classified the subject as AD or non-AD patient by training either a Logistic Regression (LR) or a Support Vector Machine (SVM) classifier. In terms of the audio modality, the authors extracted handcrafted acoustic features, i.e., ComParE, COVAREP, etc. and deep acoustic embeddings, i.e., YAMNet, VGGish, etc. Similarly to the textual modality, they used feature aggregation techniques and trained a LR and SVM classifier. Results indicated that the majority-level approach of text models yielded the highest evaluation results, while the fusion of textual and acoustic modalities led to a degredation in performance. Shah et al. \cite{10.3389/fcomp.2021.624659} introduced a weighted majority-vote ensemble meta-algorithm for classification utilizing the modalities of speech and transcripts. For the textual modality, the authors extracted language and fluency features, including the type-token ratio, the number of verbs per utterance, etc. and n-gram features. For the acoustic modality, the authors extracted four feature sets using the OpenSMILE v2.1 toolkit. After that, the authors applied dimensionality reduction techniques, i.e., Principal Component Analysis, and feature selection techniques, i.e., ANOVA F-values. Finally, shallow machine learning classifiers were trained. Best results were obtained by using only the textual modality, while the majority vote approach by combining textual and acoustic modalities led to a decrease in the classification performance. Cummins et al. \cite{cummins2020comparison} exploited also a majority level approach for detecting AD patients. For the acoustic modality, the authors exploited bag-of-audio-words, siamese networks, and end-to-end convolutional neural networks, while for the textual modality, a bidirectional Hierarchical Attention Network and a BiLSTM with an attention mechanism were used. Findings indicated that the majority vote approach achieved the highest accuracy accounting for 85.20\%. Sarawgi et al. \cite{sarawgi20_interspeech} trained acoustic and language models separately and proposed three kinds of ensemble modules for classification. Specifically, the authors experimented with hard ensemble, meaning that a majority vote was taken between the predictions of the three individual models. A soft ensemble was also proposed, where a weighted sum of the class probabilities was computed for final decision, in order to leverage the confidence of the predictions. Also, a learnt ensemble was exploited, where  a logistic regression classifier was trained using class probabilities as inputs. Results showed that the hard ensemble approach yielded the best results. Mittal et al. \cite{mittal2021multimodal} proposed a late fusion strategy using the modalities of speech and transcripts. Firstly, they trained separately acoustic and language models. For the acoustic modality, the authors trained a VGGish model with log-mel spectrograms. For the textual modality, the authors concatenated the representation obtained by BERT, Sentence-BERT, and fastText-CNN. Finally, the probabilities calculated by the audio and text-based model were combined in a weighted manner, and a threshold was fixed for classifying the persons into AD and healthy control.

There are also approaches, which add or concatenate the representation vectors of different modalities during training. However, in this way, the inherent correlations between the different modalities are not captured. On the contrary, equal importance is assigned to the different modalities. Mahajan and Baths \cite{10.3389/fnagi.2021.623607} introduced both unimodal and multimodal approaches for detecting AD patients. In terms of the multimodal approach, the introduced deep learning architecture consisted of GRU, Dense, CNN, BiLSTM, and Attention layers. For fusing the two modalities, the authors exploited the concatenation operation. Results indicated that the multimodal architecture achieved equal accuracy with the unimodal approach utilizing only text. Also, none of the proposed architectures surpassed the results of the baseline study \cite{luz20_interspeech}. Also, the authors in \cite{10.3389/fcomp.2021.624683} proposed both unimodal and multimodal approaches. With regards to the multimodal approaches, the authors used the add and concatenation operation for fusing the modalities of speech and transcripts. Best results were obtained by concatenating the representations obtained by BERT and Speech BERT. The authors in \cite{pan21c_interspeech} proposed both unimodal and multimodal approaches. Regarding unimodal approaches using speech data, the authors extracted acoustic features and trained four shallow machine learning classifiers. For the language modality, the authors trained a BERT model. In terms of the multimodal approach, the authors simply concatenated the representations obtained by BERT and acoustic modality. Results on the test set indicated that the fusion approach achieved lower performance than the unimodal one using the textual modality.

Pappagari et al. \cite{pappagari21_interspeech} trained acoustic and language models separately and used the output scores as inputs to a Logistic Regression classifier for obtaining the final prediction. For the language models, the authors used automatic speech recognition models for transcribing the recordings and employed a BERT model. For the acoustic modality, the authors used x-vectors for classifying subjects into AD patients and non-AD ones. Also, they extracted eGeMAPS, VGGish, prosody features, etc. and trained Logistic Regression and XGBoost classifiers. The authors stated that the combination of the different models and the BERT model trained on automatic transcripts achieved equal accuracy on the test set. Similarly, the authors in \cite{pappagari20_interspeech} trained also acoustic and language models separately. In terms of the acoustic models, the authors extracted the x-vectors and trained a Probabilistic Linear Discriminant Analysis classifier. For the textual modality, the authors employed a BERT model. For fusing the two modalities, the authors employed the scores from the whole training subset to train a final fusion GBR model that was used to perform the fusion of scores coming from the acoustic and transcript-based models for the challenge evaluation. Results showed that the proposed approach was the best performing one.

A different approach was proposed by \cite{rohanian20_interspeech}. More specifically, the authors extracted textual and acoustic features and passed them through two different branches of BiLSTM layers. A gating mechanism consisting of highway networks was proposed for fusing the two modalities. However, the authors did not experiment with replacing the proposed fusion method with a concatenation operation via an ablation study. Thus, this fusion method cannot guarantee performance improvement. Similarly, \cite{rohanian21_interspeech} used BERT instead of BiLSTM for extracting the text representation and stated that the BiLSTM performed better than BERT due to the fewer parameters used.

In~\cite{10.3389/fnagi.2022.830943}, the authors introduced three approaches for fusing the textual and visual modalities. Specifically, the authors passed the transcripts through BERT model and the images (log-Mel spectrogram, delta, delta-delta of the audio files) into a Vision Transformer. In terms of the first method for fusing the two modalities, the authors used a co-attention mechanism. Regarding the second method, the authors concatenated the representation matrices of the two modalities and passed the concatenated matrix through a gated-self attention layer. Finally, the authors used a Multimodal Shifting Gate, where they injected extra information to the BERT model, instead of capturing cross-modal interactions. Our work is different from \cite{10.3389/fnagi.2022.830943}, since we exploit context-based self-attention, optimal transport domain adaptation methods, Optimal Transport Kernel, DeiT, label smoothing, and one different method for fusing the representation matrices of self and cross-attention features.

The authors in \cite{9926818} exploited a Tensor Fusion Layer for explicitly aggregating unimodal, bimodal, and trimodal interactions. However, a high-dimensional output vector is generated through this method leading to the problem of overfitting and utilization of many resources.

Ilias et al. \cite{ILIAS2023101485} introduced both unimodal and multimodal approaches. In terms of the unimodal approaches, the authors converted the audio files into log-Mel spectrograms, delta, and delta-delta and exploited several pretrained models, including Vision Transformer, ResNet50, VGG16, EfficientNet-B2, etc. Regarding the multimodal approach, the authors experimented with three methods for fusing the two modalities. Firstly, they used the concatenation operation. Secondly, they employed a Gated Multimodal Unit for controlling the influence of each modality. Finally, they exploited the crossmodal attention and stated that crossmodal models outperform the competitive multimodal ones.

\subsection{Other Multimodal Tasks}

Villegas et al. \cite{sanchez-villegas-etal-2021-analyzing} introduced multimodal approaches for inferring the political ideology of an ad sponsor and identifying whether the sponsor is an official political party of a third-party organization. The authors employed BERT and EfficientNet \cite{tan2019efficientnet} for extracting textual and visual representations respectively. They concatenated these two representations and passed the resulting vector to an output layer for binary classification. Results suggested that the combination of both modalities led to a surge in the classification performance.

Villegas and Aletras \cite{sanchez-villegas-aletras-2021-point} proposed multimodal approaches for the task of point-of-interest type prediction. Specifically, the authors exploited BERT and Xception \cite{8099678} for extracting text and visual representations respectively. Next, they introduced three different architectures for fusing the two modalities. First, they exploited the Gated Multimodal Unit introduced by \cite{arevalo2020gated}. Secondly, inspired by \cite{tsai-etal-2019-multimodal}, they proposed a model for modeling the cross-modal interactions. Finally, the authors introduced an architecture, which includes the gated multimodal mechanism and the cross-attention layers on the top of the gated multimodal mechanism. Findings suggested that the proposed architecture yielded new state-of-the-art results outperforming significantly the previous text-only models.

Gu et al. \cite{gu-etal-2018-hybrid} presented a deep multimodal network with both feature attention and modality attention to classify utterance-level speech data. The authors used the modalities of audio signal and text data as input to the deep neural network. In terms of the modality fusion approach proposed, it consisted of three main parts, namely the modality attention module, the weighted operation, and the decision making module. Findings showed that the multimodal system achieved state-of-the-art performance and was tolerant to noisy data indicating in this way its generalizability. 

Pan et al. \cite{pan-etal-2020-modeling} proposed a multimodal architecture for detecting sarcasm in Twitter. More specifically, the authors exploited the ResNet-152 model and obtained a visual representation. Regarding the textual modality, they used a pretrained BERT model. After obtaining embeddings for the input sequence and the hashtags included in the sequence, the authors passed the corresponding embeddings through encoders of the transformer. For modeling the cross-modal interactions, an additional encoder was used, where the visual representation corresponded to the key and value, while the sequence representation corresponded to the query. In addition, an intra-modality attention approach was used, which gets as input the sequence and the hashtag representations. The outputs obtained were concatenated and passed to an output layer for the final prediction. Findings stated that the proposed architecture achieved state-of-the-art results.

Inspired by the transformer model in machine translation \cite{10.5555/3295222.3295349}, the authors in \cite{Yu_2019_CVPR} presented some multimodal approaches for the task of visual question answering. More specifically, the authors employed a self-attention and a guided-attention unit for capturing the intra- and inter-modal interactions respectively. Next, they obtained a Modular Co-Attention layer, which constitutes the modular composition of the self-attention and guided-attention units. Finally, the authors proposed a deep Modular Co-Attention Network consisting of cascaded Modular Co-Attention layers. Results indicated that the introduced approach surpassed the existing co-attention models.

Zadeh et al. \cite{zadeh-etal-2017-tensor} introduced a novel model, termed Tensor Fusion Network, for the task of multimodal sentiment analysis. The authors used visual, language, and acoustic modalities. For capturing the intra-modal interactions, the authors proposed three Modality Embedding Subnetworks. For capturing the inter-modal interactions, the Tensor Fusion layer has been used. Finally, the authors employed the Sentiment Inference Subnetwork, which is conditioned on the output of the Tensor Fusion layer and performs sentiment inference. Results indicated a surge in performance in comparison with existing research initiatives.

Cai et al. \cite{cai2019multi} presented a multimodal approach for sarcasm detection in Twitter. The authors used the modalities of text features, image features, and image attributes. After extracting image features and attributes, the authors leveraged attribute features and BiLSTM layers for extracting the text features. Next, the authors employed a representation fusion approach for reconstructing the features of the three modalities. Finally, they proposed a modality fusion approach motivated by \cite{gu-etal-2018-hybrid}. Results showed the effectiveness of the proposed architecture and the usefulness of the three modalities.

A different approach was proposed by \cite{Pramanick_2022_WACV}, where the authors utilized optimal transport for capturing the cross-modal interactions and self attention mechanisms for capturing the intra-modal correspondence. Specifically, they exploited three different modalities, namely visual, language, and acoustic modalities. After utilizing self-attention and optimal transport methods, they used the multimodal attention fusion method introduced by \cite{gu-etal-2018-hybrid}. Experiments conducted towards the sarcasm and humor detection tasks demonstrated valuable advantages over existing research initiatives.

Yu et al. \cite{yu2019multimodal} introduced an approach for capturing both the inter- and intra-modal interactions for the visual question answering and the visual grounding tasks using the modalities of text and image. Specifically, after obtaining text and visual representations, they passed these two representations through a unified attention block. The authors proposed also a variation of the self-attention mechanism by introducing a novel gating model. Findings showed the effectiveness of the proposed approach on five datasets. 

\subsection{Related Work Review Findings}

From the research works mentioned above, it is obvious that despite the fusion methods, which have been proposed for many tasks, little work has been done in terms of the multimodal approaches for the dementia detection task. More specifically, for the task of dementia detection, existing research initiatives employ majority-vote approaches or propose early and late fusion strategies. Applying majority-vote approaches and late fusion strategies entails significant increase in the computational time, since multiple models must be trained separately. For early fusion approaches, features obtained by different modalities are concatenated at the input level. All these approaches do not capture either the inter- or the intra-modal interactions effectively. Recently, there have been proposed some studies trying to address the limitations of fusing the different modalities. However, limitations still exist. Finally, no study has addressed the problem of creating overconfident models, which is crucial in health-related tasks.

Therefore, the present study differs from the existing state-of-the-art approaches, since we \textit{(i)} exploit DeiT for extracting a visual representation, \textit{(ii)} use optimal transport domain adaptation methods for capturing the inter-modal interactions and a self-attention mechanism with a gating model for capturing the intra-modal interactions, \textit{(iii)} exploit a context-based self-attention mechanism, where we enhance the self-attention layer with contextual information by using three approaches, \textit{(iv)} employ label smoothing for calibrating our proposed approaches, and \textit{(v)} introduce a new method in the task of dementia recognition from spontaneous speech for fusing the self and cross-attention features.

\section{Data \& Task} \label{Dataset}

\subsection{ADReSS Challenge Dataset}

We use the ADReSS Challenge Dataset \cite{luz20_interspeech} for conducting our experiments. The data corresponds to spoken picture descriptions elicited from participants through the Cookie Theft picture from the Boston Diagnostic Aphasia Exam \cite{10.1001/archneur.1994.00540180063015}. We choose the specific dataset, since it minimizes several kinds of biases, which could influence the validity of the proposed approaches during the training and evaluation procedure. Specifically, in contrast to other datasets, the ADReSS Challenge dataset is matched for gender and age. In addition, it is balanced, since it includes 78 AD and 78-non AD patients. What is also worth noting is the fact that the ADReSS Challenge dataset has been carefully selected so as to mitigate common biases often overlooked in evaluations of AD detection methods, including repeated occurrences of speech from the same participant (common in longitudinal datasets) and variations in audio quality. To be more precise, recordings have been acoustically enhanced with stationary noise removal and audio volume normalization has been applied across all speech segments to control for variation caused by recording conditions, such as microphone placement. The ADReSS Challenge dataset has been divided by the organizers into a train and a test set. The train set consists of 54 AD and 54 non-AD patients, while the test set comprises 24 AD patients and 24 non-AD ones.

\subsection{ADReSSo Challenge Dataset}

To further verify the effectiveness of our proposed approaches, we use the ADReSSo Challenge Dataset \cite{luz21_interspeech}. Similar to the ADReSS Challenge Dataset, this dataset has been created in a way for eliminating various kinds of biases. Additionally, each participant is asked to describe the Cookie Theft picture from the Boston Diagnostic Aphasia Examination. It is divided into a train and a test set. The train set consists of 87 AD patients and 79 non-AD ones, while the test set includes 35 AD and 36 non-AD patients. Also, this dataset includes only audio files. No transcripts are provided. Therefore, one should convert the speech into text automatically via Automatic Speech Recognition (ASR) methods. Specifically, we use whisper\footnote{https://github.com/openai/whisper} \cite{radford2022robust} and get the automatically generated transcripts per audio file.

\subsection{Task}

Let a labeled dataset consist of audio files and their corresponding transcripts. Each transcript along with its audio file belongs to an AD patient or non-AD patient. The task is to identify if a specific transcript along with its audio file corresponds to an AD patient or to a person belonging to the healthy control group (binary classification problem).

\section{Predictive Models} \label{predictive_models}

\subsection{Architecture}

In this section, we describe our proposed deep learning architectures for detecting AD patients. The proposed architectures are illustrated in Fig.~\ref{proposed_architectures}. Due to the fact that the manual transcripts have been annotated using the CHAT coding system \cite{10.1162/coli.2000.26.4.657}, we use the PyLangAcq library \cite{lee-et-al-pylangacq:2016} for having access to these transcripts. In addition, we use the Python library, called librosa \cite{brian_mcfee_2022_6759664,mcfee2015librosa}, and convert each audio file into a log-Mel spectrogram, its delta, and delta-delta. In this way, we create an image consisting of three channels. For all the experiments conducted, we use 224 Mel bands, hop length equal to 1024, and a Hanning window. Each image is resized to $\left (224 \times 224 \right)$ pixels.

Firstly, we pass each transcript through a BERT \cite{devlin-etal-2019-bert} model and the corresponding image through a DeiT \cite{pmlr-v139-touvron21a} model. Formally, let $X \in \mathbb{R}^{n \times D}$ and $Y \in \mathbb{R}^{T \times D}$ be the outputs of the BERT and DeiT pretrained models respectively. Next, we pass $Y$ through an Optimal Transport Kernel introduced by \cite{mialon2021a}, in order to ensure that the sequence length of $Y$ is equal to the sequence length of $X$, i.e., $T=n$. Let $S \in \mathbb{R}^{T \times D}$, where $T=n$, denote the output representation of the Optimal Transport Kernel. 

\begin{figure*}[!htb]
\centering
\begin{subfigure}[t]{0.95\columnwidth}
\includegraphics[width=\linewidth]{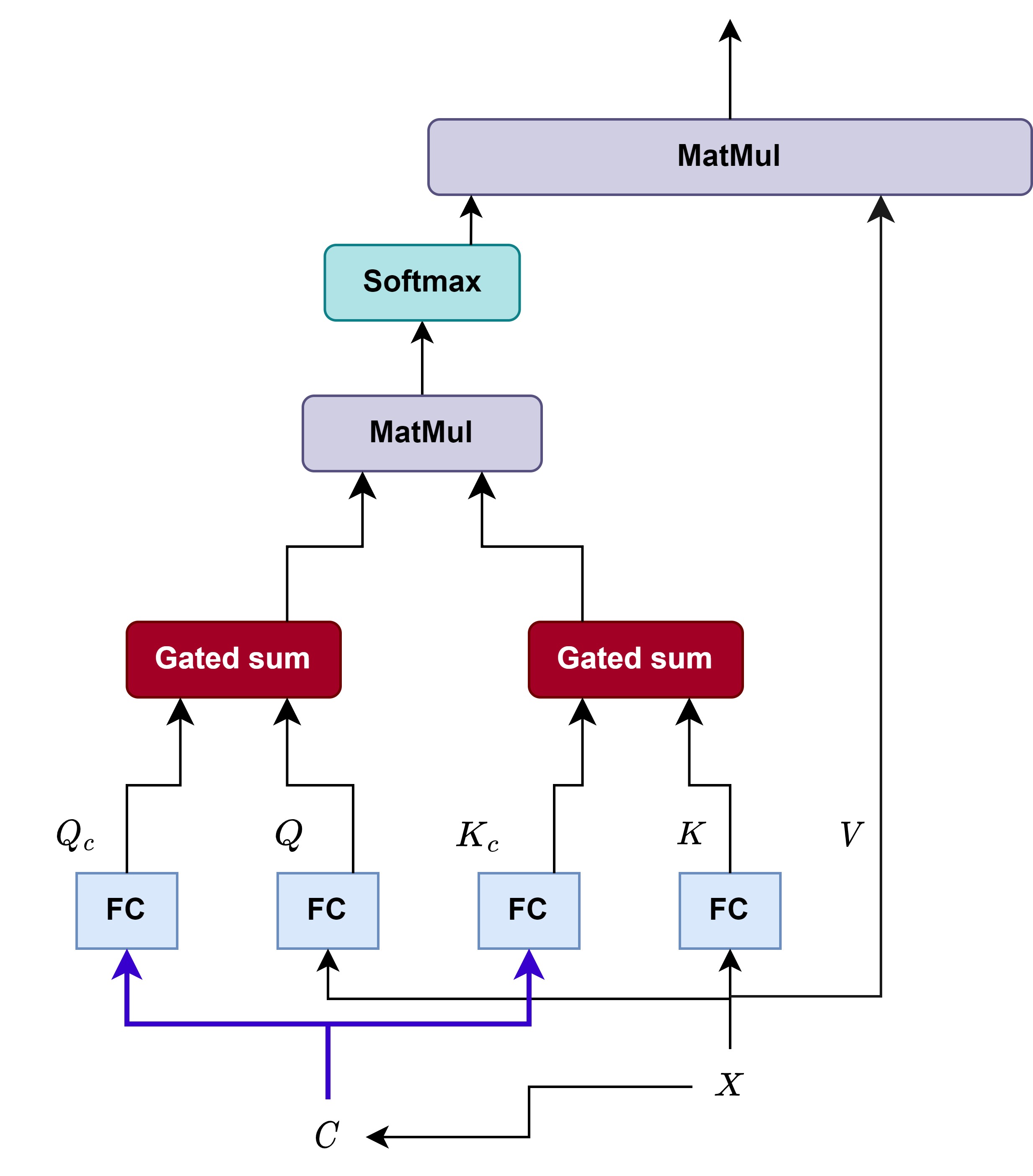}
\caption{Context-based Self-Attention. This method is different from the conventional self-attention mechanism, since it exploits a contextual information vector $C$.}
\label{cbsa}
\end{subfigure}
\begin{subfigure}[t]{0.95\columnwidth}
\centering
\includegraphics[width=\linewidth]{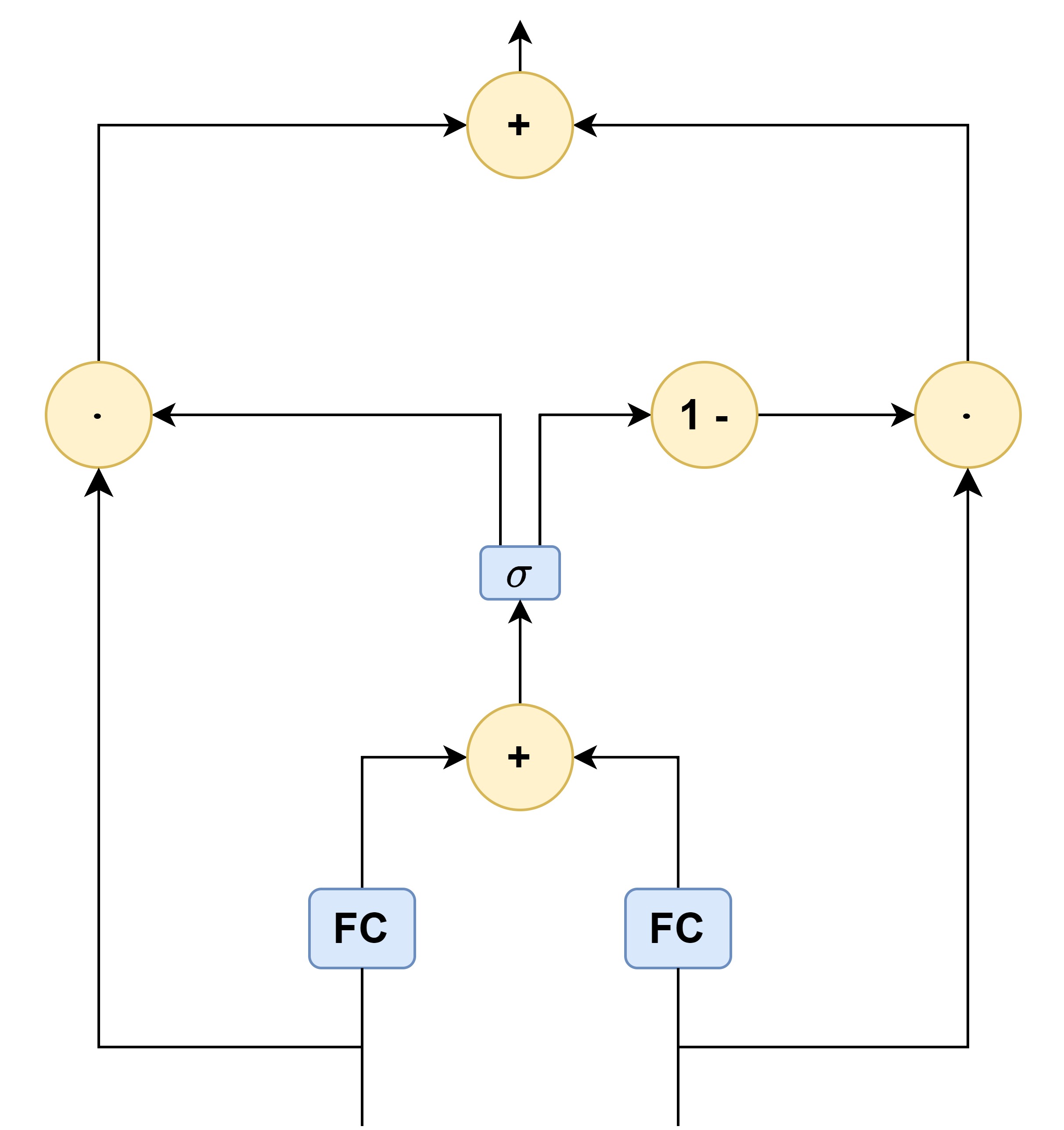}
\caption{Gated sum. This unit is used for quantifying the contribution of the original representation $X$ and the context vector $C$ to the attention weight prediction.}
\label{gated_sum}
\end{subfigure}
\caption{Context-based Self-Attention}
\label{context_based_self_attention}
\end{figure*}

\begin{figure*}[!htb]
\centering
\begin{subfigure}[t]{0.95\columnwidth}
\includegraphics[width=\linewidth]{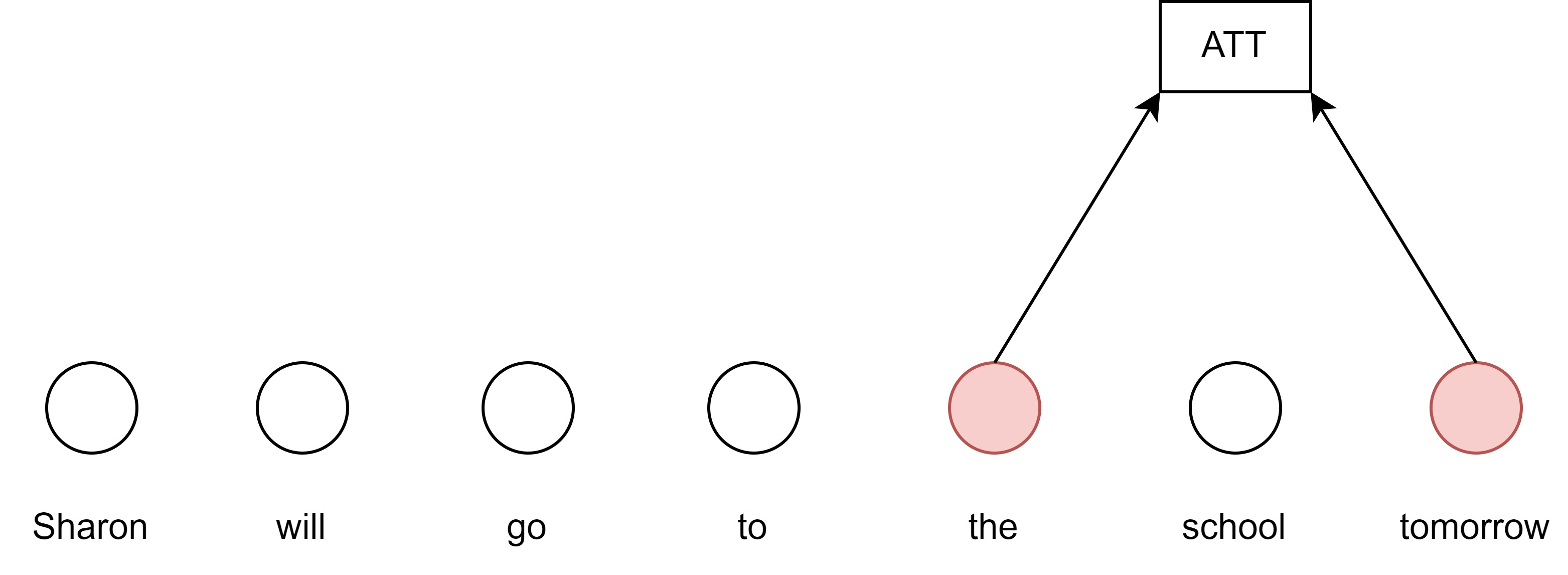}
\caption{Conventional Self-Attention. This method calculates the attention weight of two items ignoring the contextual information.}
\label{conventional}
\end{subfigure}
\begin{subfigure}[t]{0.95\columnwidth}
\centering
\includegraphics[width=\linewidth]{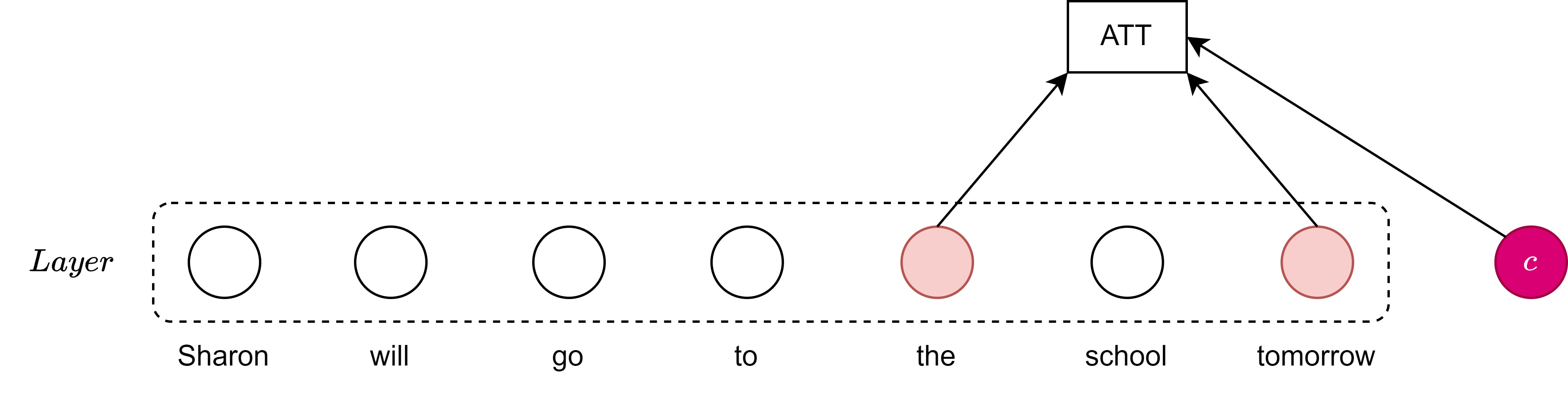}
\caption{Global Context. This method captures the summary representation of the input sentence through an average operation.}
\label{global_context}
\end{subfigure}

\begin{subfigure}[t]{0.95\columnwidth}
\centering
\includegraphics[width=\linewidth]{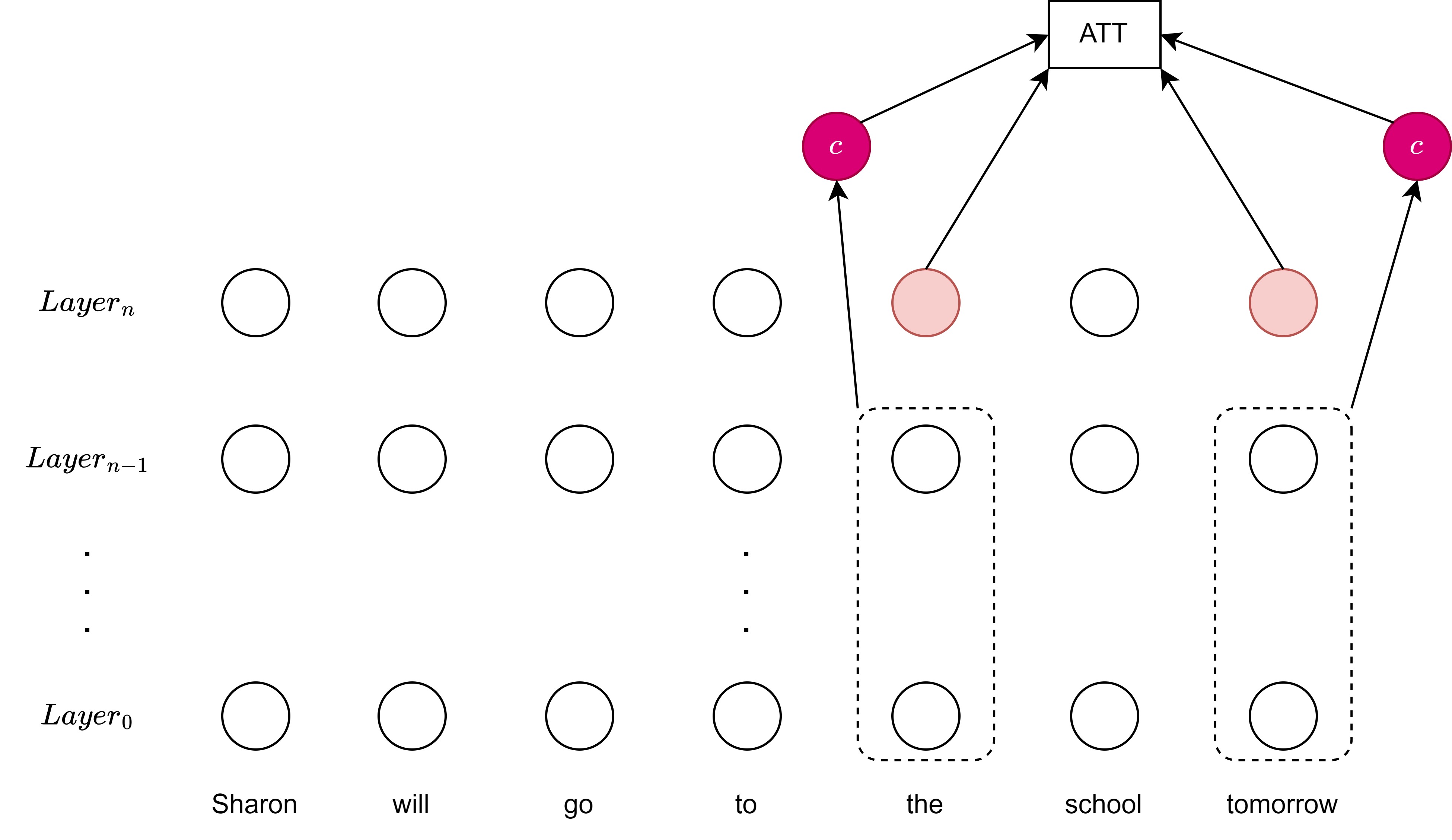}
\caption{Deep Context. This method captures both the low- and high-level syntactic and semantic information.}
\label{deep_context}
\end{subfigure}
\begin{subfigure}[t]{0.95\columnwidth}
\centering
\includegraphics[width=\linewidth]{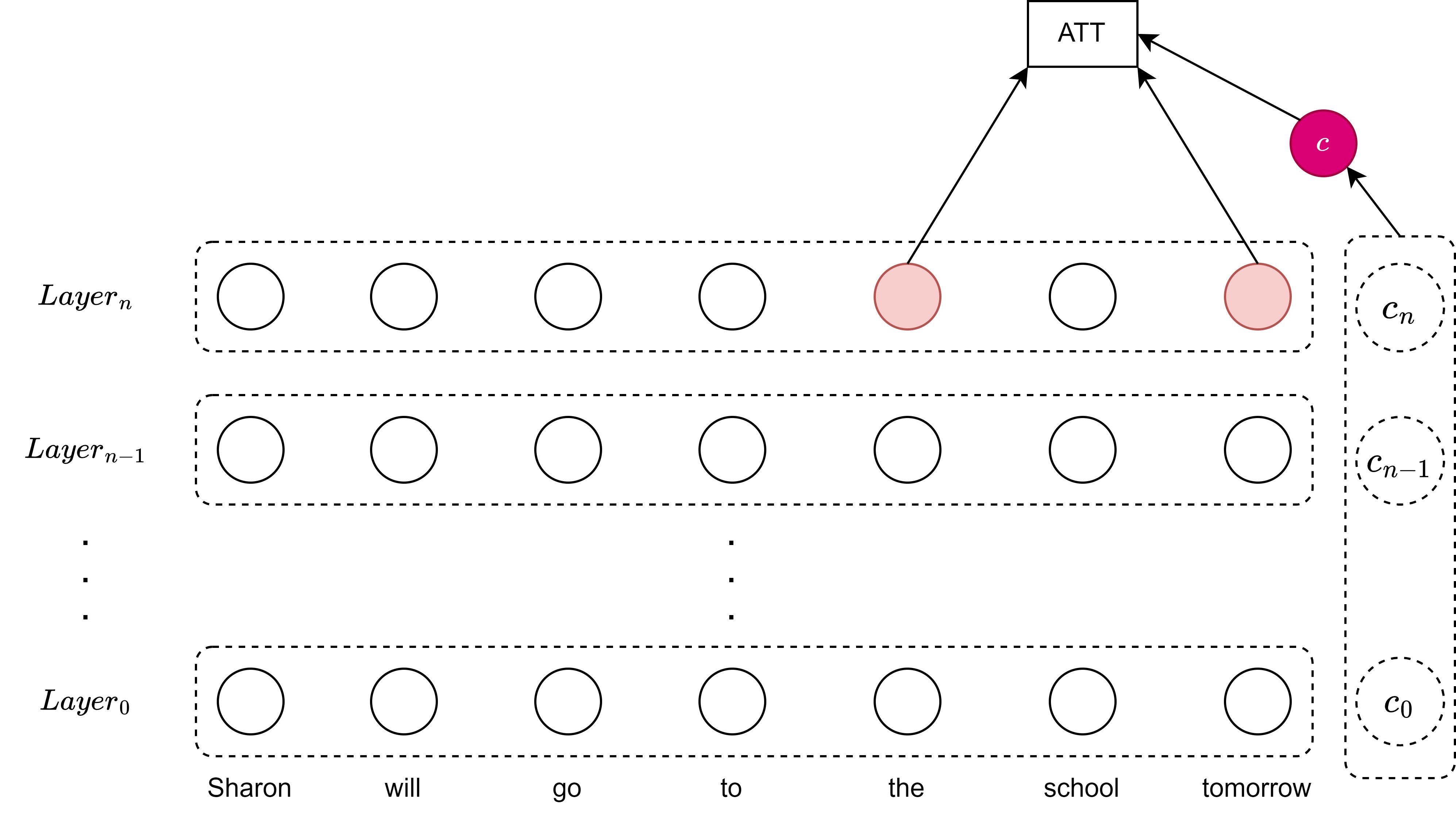}
\caption{Deep-Global Context. This method combines the concepts of global and deep context.}
\label{deep_global}
\end{subfigure}
\caption{Self-Attention based on different context-vectors}
\label{newwww}
\end{figure*}

\noindent \textbf{Context-Aware Self Attention for the textual modality:} Fig.~\ref{conventional} illustrates the conventional self-attention mechanism, which individually calculates the attention weight of two items, i.e., "the" and "tomorrow", ignoring the contextual information. In this study, we aim to enhance the self-attention layer by adding contextual information. Therefore, we exploit the context-based self-attention layer \cite{CHEN2022108980}, which is illustrated in Fig.~\ref{context_based_self_attention}. We observe that this layer receives as input the input sequence denoted by $X$ and the contextual information vector denoted by $C$.

We transform the input sequence $X$ into a query, key, and value matrix, as described via the Equation \ref{query_x}:

\begin{equation}
    Q = XW_q, K = XW_k, V = X
    \label{query_x}
\end{equation}
, where $W_q  \in \mathbb{R}^{D \times D_q}, W_k \in \mathbb{R}^{D \times D_k}$ are learnable weight matrices.

As described in Equations~\ref{query_C} and \ref{query_K}, the context vector $C \in \mathbb{R}^{n \times D_c}$ is transformed to a contextual query matrix $Q_c \in \mathbb{R}^{n \times D_q}$ and a contextual key matrix $K_c \in \mathbb{R}^{n \times D_k}$:

\begin{equation}
    Q_c = CW_q ^c
    \label{query_C}
\end{equation}
, where $W_q ^c \in \mathbb{R}^{D_c \times D_q}$ is a learnable weight matrix.
\begin{equation}
    K_c = CW_k ^c
    \label{query_K}
\end{equation}
, where $W_k ^c \in \mathbb{R}^{D_c \times D_k}$ is a learnable weight matrix.

Next, we exploit gated sum, as illustrated in Fig.~\ref{gated_sum}, for quantifying the contribution of the input sequence $X$ and the contextual vector $C$ to the attention weight prediction. Finally, we get new query and key matrices denoted by $\overline{\rm Q} \in \mathbb{R}^{n \times D_q}$ and $\overline{\rm K} \in \mathbb{R}^{n \times D_k}$ respectively. We describe the equations governing the gated sum below:

\begin{equation}
    g_q = \sigma \left(QW_g ^Q + Q_c W_g ^{Q_c} \right)
\end{equation}
, where $W_g ^Q, W_g ^{Q_c} \in \mathbb{R}^{D_q \times 1}$ are learnable parameters.
\begin{equation}
    g_k = \sigma \left(KW_g ^K + K_c W_g ^{K_c} \right)
\end{equation}
where $W_g ^K, W_g ^{K_c} \in \mathbb{R}^{D_k \times 1}$ are learnable parameters.

$q_q$ and $g_k$ indicate the weight of the importance of the contextual information. 
\begin{equation}
    \overline{\rm Q} = (1-g_q)Q +g_q Q_c
\end{equation}

\begin{equation}
    \overline{\rm K} = (1-g_k)K +g_k K_c
\end{equation}
Therefore, we obtain new query and key matrices. Finally, we calculate the self-attention via the equation mentioned below:

\begin{equation}
    Attention(\overline{\rm Q}, \overline{\rm K}, V) = softmax \left(\frac{\overline{\rm Q} \cdot \overline{\rm K}^T}{\sqrt{D_k}}\right) V
\end{equation}

Next, we describe three methods, namely Global Context, Deep Context, and Deep-Global Context, for calculating the contextual vector $C$. Specifically, we follow \cite{CHEN2022108980,Yang_Li_Wong_Chao_Wang_Tu_2019} to represent the context vector ($C$), which is composed of internal representation.

\begin{itemize}
    \item \textbf{Global Context:} Fig.~\ref{global_context} illustrates the global context strategy. More specifically, the global context indicates the mean operation over the input sequence for summarizing the input representation. Let $X=[x_1,x_2,...,x_n] \in \mathbb{R}^{n \times D}$. We calculate the context representation $C$ as defined in Eq.~\ref{context_vector_global_context}. Note that the output of Eq.~\ref{context_vector_global_context} is a vector, i.e., $C \in \mathbb{R}^D$, instead of a matrix. To facilitate subsequent calculation operations, we use Eq.~\ref{stack_global}, where we obtain the contextual matrix $\mathbf{C} \in \mathbb{R}^{n \times D}$.

    \begin{equation}
        C = \overline{\rm X} = Avgpool(X)=\frac{1}{n} \sum_{1}^{n} x_i
        \label{context_vector_global_context}
    \end{equation}

    \begin{equation}
        \mathbf{C} = stack\left (C,C,...,C \right)
        \label{stack_global}
    \end{equation}
    \item \textbf{Deep Context:} By deeply stacking self-attention layers, the model captures only high-level syntactic and semantic information neglecting the lower-level information. Therefore, as shown in Fig.~\ref{deep_context}, the deep context strategy enables the layer to fuse different types of syntactic and semantic information captured by different layers.
    
    Formally, taking $X=[x_1,x_2,...,x_n] \in \mathbb{R}^{n \times D}$ as the initial input sequence $X^0$, and the output of the $L^{th}$ layer is $X^l=[x_1 ^l,x_2 ^l,...,x_n ^l] \in \mathbb{R}^{n \times D}$, the deep context matrix $C \in \mathbb{R}^{n \times D}$ can be represented as follows:
    \begin{equation}
        C = \hat{X} W_c ^0
    \end{equation}
    \begin{equation}
        \hat{X}=concat(X^0,X^1,...,X^{l-1}) \in \mathbb{R}^{n \times lD}
    \end{equation}
    , where $W_c ^0 \in \mathbb{R}^{lD \times D}$ is a learnable parameter matrix. \textit{concat(.)} denotes join operation. 
    \item \textbf{Deep-Global Context:} The deep-global context strategy combines the strategies of global context and deep context as described before. The deep-global context strategy is illustrated in Fig.~\ref{deep_global} and is described via the equations below:
    
    \begin{equation}
        C = \overline{\rm C} W_{\overline{\rm C}}^0
        \label{C_deep_global}
    \end{equation}
    \begin{equation}
        \overline{\rm C}=concat(C^0,C^1,...,C^{l-1})
    \end{equation}
    , where $C^j = Avgpool(X^j), C^j \in \mathbb{R}^D$. Therefore, $\overline{\rm C} \in \mathbb{R}^{lD}$.
    In addition, $W_{\overline{\rm C}}^0 \in \mathbb{R}^{lD \times D}$. Thus, we obtain $C$ of Eq.~\ref{C_deep_global}, as $C \in \mathbb{R}^{D}$. 
    
    As mentioned before, to facilitate subsequent calculation operations, matrix $\mathbf{C} \in \mathbb{R}^{n \times D}$ is obtained through the stack operation, as follows: $\mathbf{C}=stack(C,C,...,C)$.
\end{itemize}

Let $F$ be the output of the context-based self-attention mechanism corresponding to the textual modality denoted by $X$. 

\begin{figure}[!htb]
\centering
\includegraphics[width=0.3\textwidth]{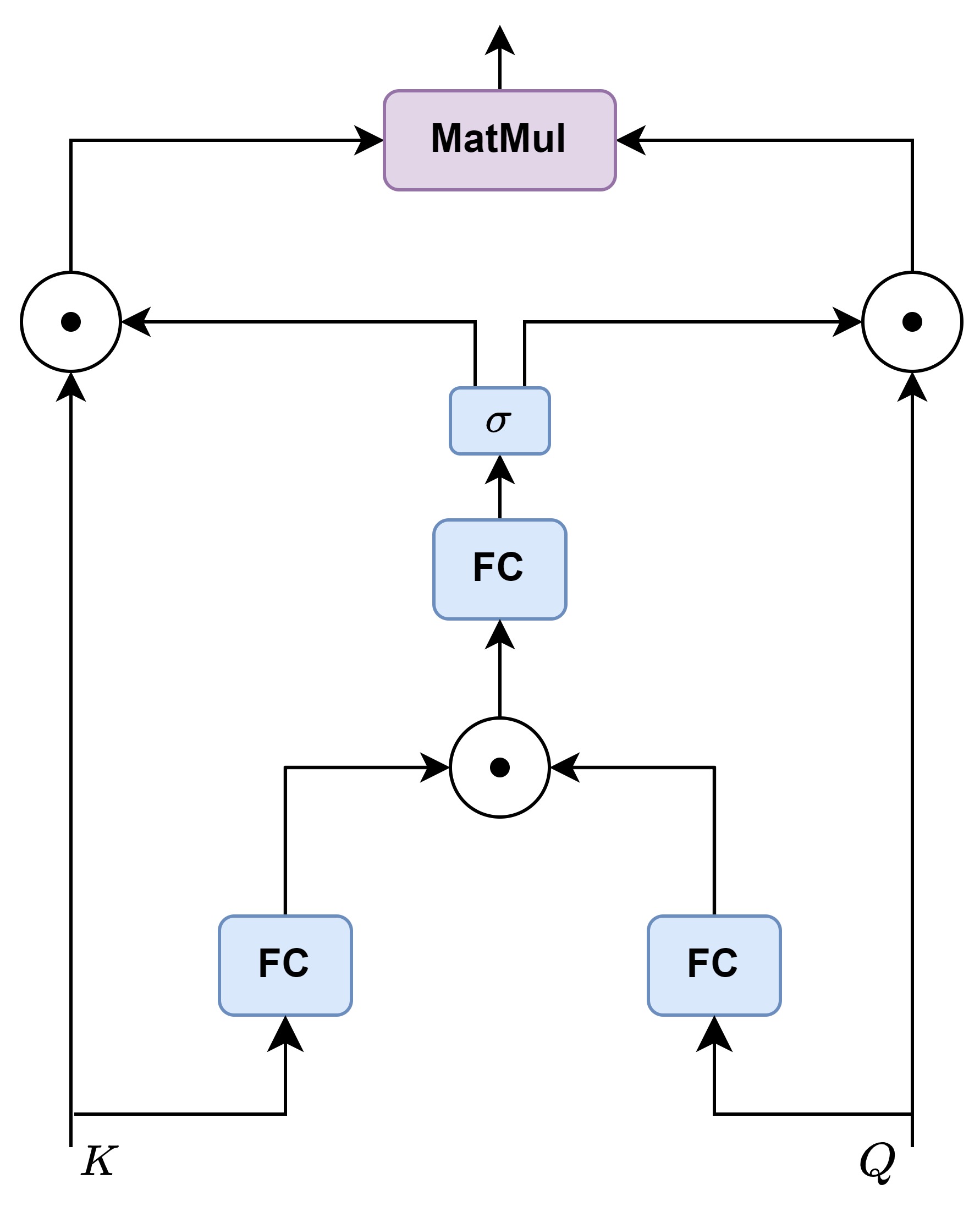}
\caption{Gated Dot-product. This gating model is incorporated in the conventional self-attention mechanism for improving the quality of the learned attention. This method is based on low-rank bilinear pooling.}
\label{gated_self_attention}
\end{figure}

\noindent \textbf{Gated Self-Attention for the image modality:} Motivated by the work of \cite{yu2019multimodal}, we pass $S$ through a self-attention mechanism, which incorporates a novel gating model, for capturing the intra-modal interactions. This gated self-attention mechanism is illustrated in Fig.~\ref{gated_self_attention}. The self-attention mechanism including the gating model is described via the equations below:

\begin{equation}
    Q = S, K = S, V = S 
    \label{first_equation}
\end{equation}

\begin{equation}
    M = \sigma \left (FC^g \left(FC^g _q \left(Q\right)\odot FC^g _k \left(K\right)\right)\right )
    \label{gating_model}
\end{equation}
where $FC^g _q, FC^g _k \in \mathbb{R}^{D \times d_g}$, $FC^g \in \mathbb{R}^{d_g \times 2}$ are three fully-connected layers, and $d_g$ denotes the dimensionality of the projected space and is equal to 64 units. $\odot$ denotes the element-wise product function and $\sigma$ the sigmoid function. In addition, $M \in \mathbb{R}^{T \times 2}$ corresponds to the two masks $M_q \in \mathbb{R}^T$ and $M_k \in \mathbb{R}^T$ for the features $Q$ and $K$ respectively.

Next, the two masks $M$ and $K$ are tiled to $\Tilde{M_q}, \Tilde{M_k} \in \mathbb{R}^{T \times D}$ and then used for computing the attention map as following:
\begin{equation}
    A^g = softmax \left (\frac{\left(Q \odot \Tilde{M_q}\right)\left(K \odot \Tilde{M_k}\right)^T}{\sqrt{D}} \right )
    \label{eq_gating_model}
\end{equation}

\begin{equation}
    H = A^g V
    \label{last_equation}
\end{equation}

Let $H$ be the output of the self-attention mechanism corresponding to the visual modality denoted by $S$. 

\noindent \textbf{Optimal Transport:} Next, we use optimal transport-based domain adaptation methods \cite{villani2008optimal,7586038,ferradans2014regularized}, i.e., Earth Mover's Distance (EMD) Transport, for transporting between each pair of modalities, which can be interpreted as domain adaptation across two modalities. Formally:

\begin{equation}
    X' = OT(S \to X)
\end{equation}
\begin{equation}
    S' = OT(X \to S)
\end{equation}

\noindent \textbf{Concatenation:} After that, we concatenate transported and self-attended features as follows:
\begin{equation}
    C = [F,X']
    \label{concat_text}
\end{equation}
\begin{equation}
    S = [H,S']
    \label{concat_image}
\end{equation}

\noindent \textbf{Fusion:} Next, we describe two methods for fusing $C$ and $S$:
\begin{itemize}
    \item \textbf{(i) Co-Attention Mechanism:} We exploit the fusion method proposed by \cite{10.5555/3157096.3157129} and implemented in prior work \cite{10.3389/fnagi.2022.830943}. Specifically, given $\left(\textbf{C} \in \mathbb{R}^{d' \times n}\right)$ and $\left(\textbf{S} \in \mathbb{R}^{d' \times T}\right)$, where $d'=2 \cdot D$, the affinity matrix $F \in \mathbb{R}^{n \times T}$ is calculated using the equation presented below:

\begin{equation}
    F = \tanh \left(C^T W_l S \right)
    \label{equation1}
\end{equation}

\noindent where $W_l \in \mathbb{R}^{d' \times d'}$ is a matrix of learnable parameters. By treating the affinity matrix as a feature, we learn to predict the attention maps via the following,

\begin{equation}
    H^s = \tanh \left(W_s S + \left(W_c C \right)F \right)
\end{equation} 
\begin{equation}
    H^c = \tanh \left(W_c C + \left(W_s S \right)F^T \right)
\end{equation}

\noindent where $W_s, W_c \in \mathbb{R}^{k \times d'}$ are matrices of learnable parameters. We set $k$ equal to 40. Then, we generate the attention weights through the softmax function as follows,

\begin{equation}
    a^s = softmax \left(w_{hs} ^T H^s \right) 
\end{equation}
\begin{equation}
    a^c = softmax \left(w_{hc} ^T H^c \right)
\end{equation}

\noindent where $a_s \in \mathbb{R}^{1 \times T}$ and $a_c \in \mathbb{R}^{1 \times n}$. $w_{hs}, w_{hc} \in \mathbb{R}^{k \times 1}$ are the weight parameters. Based on the above attention weights, the attention vectors are obtained via the following equations:
\begin{equation}
    \hat{s} = \sum_{i=1}^{T} a_i ^s s^i, \quad \hat{c}=\sum_{j=1}^{n} a_j ^c c^j
\end{equation}

\noindent where $\hat{s} \in \mathbb{R}^{1 \times d'}$ and $\hat{c} \in \mathbb{R}^{1 \times d'}$.

Finally, these vectors are concatenated $p = [\hat{c}, \hat{s}]$. We apply a dropout layer with a rate of 0.5. Then, this vector is passed through a Dense Layer consisting of 128 units with a ReLU activation function. We apply also a dropout layer with a rate of 0.2. Finally, we use a dense layer consisting of two units, which gives the final output.

The proposed architecture is illustrated in Fig. \ref{Optimal_Transport_CoAttention}.

\item \textbf{(ii) Attention-based fusion:} Motivated by the work of \cite{Yu_2019_CVPR}, we design an attentional reduction model for $C$, as defined in Equation \ref{concat_text} (or $S$, as defined in Equation \ref{concat_image}), for obtaining its attended feature $\Tilde{c}$ (or $\Tilde{s}$). To the best of our knowledge, this is the first study utilizing this fusion method in the task of dementia detection from spontaneous speech. Taking $C$ as an example, we describe the attention reduction model used in this study via the equations presented below:

\begin{equation}
    \alpha^c = softmax \left(MLP \left(C \right) \right)
\end{equation}
, where $\alpha^c$ refers to the learned attention weights and \textit{MLP} is given by the equation below:
\begin{equation}
    MLP = FC(128) - ReLU - Dropout(0.1) - FC(1)
\end{equation}
\begin{equation}
    \Tilde{c} = \sum_{i=1}^{n} \alpha_i ^c c_i
\end{equation}
, where we obtain the attended feature $\Tilde{c}$ for $C$.

We obtain the attended feature $\Tilde{s}$ using an independent attention reduction model in the same way. Having computed $\Tilde{c}$ and $\Tilde{s}$, we design the linear multimodal fusion function as follows:
\begin{equation}
    z = LayerNorm \left(W_c ^T \Tilde{c} + W_s ^T \Tilde{s} \right)
\end{equation}
, where $W_c, W_s \in \mathbb{R}^{d' \times d_z}$ are two linear projection matrices, $d_z$ is the common dimensionality of the fused feature and is equal to 128, and LayerNorm \cite{ba2016layer} is used for stabilizing the training.
Finally, we pass $z$ to a dense layer consisting of two units, which gives the final prediction.

The proposed architecture is illustrated in Fig. \ref{Optimal_Transport_Attention_Based_Fusion}.

\begin{figure*}[!htb]
\centering
\begin{subfigure}[t]{1\textwidth}
\includegraphics[width=1\linewidth]{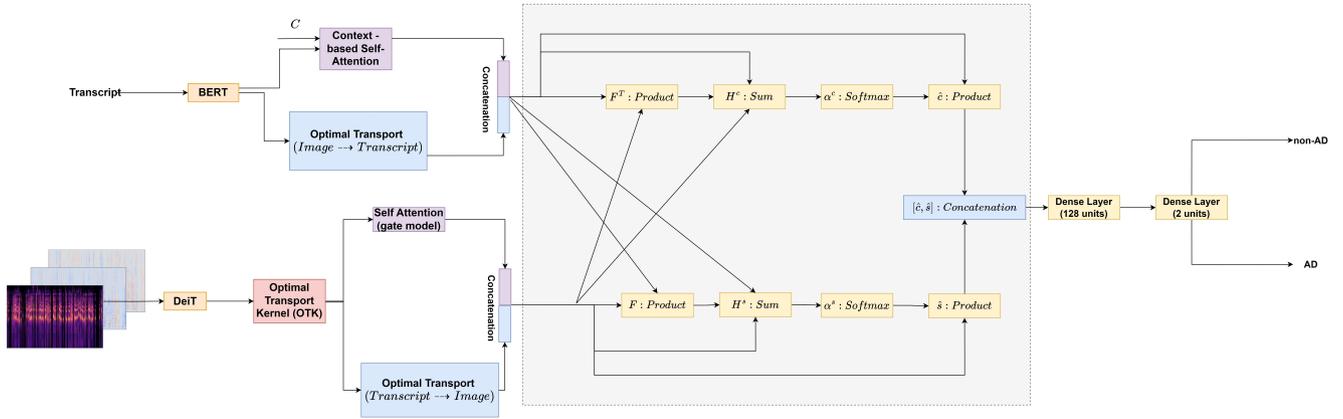}
\caption{Co-Attention. The shaded box corresponds to the co-attention mechanism. This method attends to the different representations simultaneously.}
\label{Optimal_Transport_CoAttention}
\end{subfigure}

\begin{subfigure}[t]{1\textwidth}
\centering
\includegraphics[width=1\linewidth]{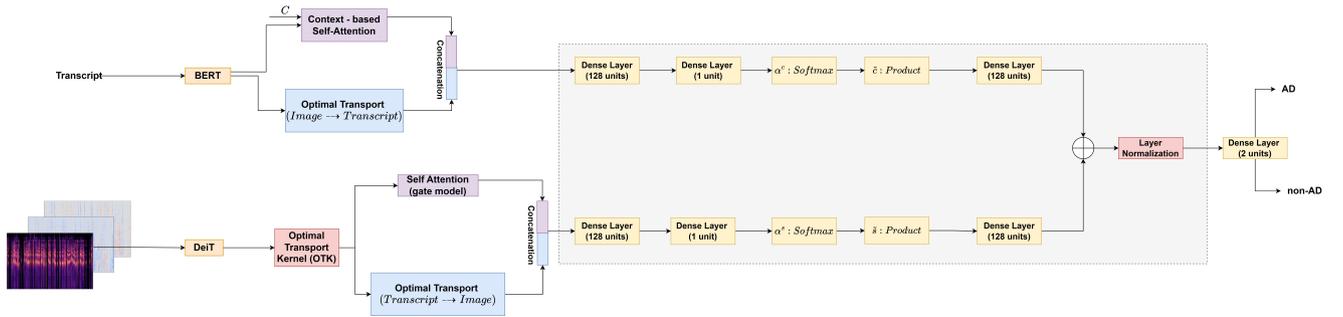}
\caption{Attention-based Fusion. The shaded box shows this fusion method. This method exploits two independent attentional reduction models. Features are fused through an add operation, while a layer normalization is used for stabilizing training.}
\label{Optimal_Transport_Attention_Based_Fusion}
\end{subfigure}
\caption{Illustration of our Proposed Architectures. For the textual modality, we use BERT, while for the image modality, we use DeiT and exploit an Optimal Transport Kernel. Next, we use optimal transport domain adaptation methods for transporting between each pair of modalities. Also, we pass the textual representation through context-based self-attention layers, while the image representation is passed through a gated self-attention layer. Finally, methods for fusing the self- and cross-attention features are presented, namely Co-Attention and Attention-based Fusion. Each shaded box shows the fusion method used, namely Co-Attention and Attention-based Fusion.}
\label{proposed_architectures}
\end{figure*}

\end{itemize}

\subsection{Model Calibration}

To prevent the model becoming too overconfident, we use label smoothing \cite{7780677,NEURIPS2019_f1748d6b}. Specifically, label smoothing calibrates learned models so that the confidences of their predictions are more  aligned with the accuracies of their predictions. 

For a network trained with hard targets, the cross-entropy loss is minimized between the true targets $y_k$ and the network's outputs $p_k$, as in $H(y,p)=\sum_{k=1}^K -y_k log(p_k)$, where $y_k$ is "1" for the correct class and "0" for the other. For a network trained with label smoothing, we modify the true targets $y_k$ to $y_k ^{LS_u}$ as shown in Eq.~\ref{label_smoothing}:

\begin{equation}
    y_k ^{LS_u} = y_k \cdot (1-\alpha)+ \frac{\alpha}{K}
    \label{label_smoothing}
\end{equation}

, where $\alpha$ is the smoothing parameter and $K$ is the number of classes.

Finally, we minimize the cross-entropy between the modified targets $y_k ^{LS_u}$ and the network's outputs $p_k$, as shown in Eq.~\ref{new_cross_entropy}:

\begin{equation}
    H(y,p)=\sum_{k=1}^{K} -y_k ^{LS_u} \cdot \log \left(p_k\right)
    \label{new_cross_entropy}
\end{equation}

\section{Experiments}

\subsection{Baselines}

We compare our introduced approaches with the following research works reported in Tables \ref{baselines_adress_challenge} and \ref{baselines_adressO_challenge}, since these research works have conducted their experiments on the ADReSS and ADReSSo test set. Specifically, Table~\ref{baselines_adress_challenge} describes the baselines used in terms of the ADReSS Challenge dataset, while Table~\ref{baselines_adressO_challenge} reports the baselines used regarding the ADReSSo Challenge dataset. Regarding Table~\ref{baselines_adress_challenge}, we are using existing published results for all the baselines except for \textit{Introduced Approaches without Label Smoothing}. In terms of Table~\ref{baselines_adressO_challenge}, we are using existing published results for all the baselines except for: (i) \textit{BERT}, and (ii) \textit{Introduced Approaches without Label Smoothing}.

\begin{table*}
\caption{Baselines (ADReSS Challenge Dataset).}
\begin{tabularx}{\textwidth}{>{\hsize=.32\hsize}X>{\hsize=.68\hsize}X}
\toprule
\multicolumn{1}{c}{\textbf{Reference/Architecture}} & \multicolumn{1}{c}{\textbf{Features/Methodology}} \\ \midrule
\multicolumn{2}{>{\columncolor[gray]{.8}}l}{\textbf{Baselines - Unimodal state-of-the-art approaches (only transcripts)}} \\
BERT \cite{9769980} &  Fine-tune a BERT model \\ \hline
\multicolumn{2}{>{\columncolor[gray]{.8}}l}{\textbf{Baselines - Multimodal state-of-the-art approaches}} \\
Fusion Maj. (3-best) \small\citep{cummins2020comparison} &  Majority Vote of the BoAW-MFCC-C125, ZFF, and bi-LSTM-Att \\ \hline
System 3: Phonemes and Audio \small\citep{edwards20_interspeech} &  Acoustic features (emobase, eGeMAPS, ComParE2016) along with feature selection techniques, transcription of the segmented text into phoneme written pronunciation using CMUDict \\ \hline
Fusion of System \small\citep{pompili20_interspeech} &  merged the x-vectors features set with the combination of linguistic feature sets (GMax/LSTM-RNNs/LSTM-RNNs-Pos) and trained a SVM classifier \\ \hline
Bimodal Network (Ensembled Output) \small\citep{koo20_interspeech} &  For the acoustic modality, the authors use VGGish, while for the textual modality, the authors exploit GloVe, Transformer-XL, POS and HC features. Finally, the authors combine the results of the models via an ensemble approach.  \\ \hline
GFI, NUW, Duration, Character 4-grams, Suffixes, POS tag, UD \small\citep{martinc20_interspeech} &  feature extraction, early fusion approach, train a Logistic Regression Classifier\\ \hline
Acoustic \& Transcript \small\citep{pappagari20_interspeech}&  For transcripts, the authors exploited BERT, while for speech, the authors used x-vector PCA-transformed coefficients.  \\ \hline
Dual BERT \small\citep{10.3389/fcomp.2021.624683} &  concatenation of the representations obtained by BERT and Speech BERT \\ \hline
Model C \small\citep{10.3389/fnagi.2021.623607} &  The authors extracted emobase, eGeMAPS, ComParE features. For the text modality, the used GloVe embeddings and pos-tags. Finally, they trained a Neural network consisting of CNN, BiLSTM, Attention, GRU, and Dense layers.  \\ \hline
Majority vote (NLP + Acoustic) \small\citep{10.3389/fcomp.2021.624659}  &  The authors extracted a set of acoustic and linguistic features. After training shallow machine learning classifiers, they chose the three best-performing acoustic models along with the best-performing language model, and computed a final prediction by taking a linear weighted combination of the individual model predictions.  \\ \hline
Audio + Text \small\citep{9459113} & majority level approach of six models \\ \hline
LSTM with Gating (Acoustic + Lexical + Dis) \small\citep{rohanian20_interspeech} &  Acoustic, Linguistic Features, Bi-LSTM, feed-forward highway layers with gating units \\ \hline
Ensemble \small\citep{sarawgi20_interspeech} &  A majority vote was taken between the predictions
of the three individual models. Specifically, the authors extracted three sets of features, namely disfluency, acoustic, and interventions, and trained three deep neural networks. \\ \hline
BERT+ViT (log-Mel spectrogram) \cite{ILIAS2023101485} & coversion of an audio file into an image of three channels, BERT for the text representation, Vision Transformer for the image representation, concatenation  \\ \hline
BERT+ViT+Gated Multimodal Unit (log-Mel spectrogram) \cite{ILIAS2023101485} & Gated Multimodal Unit to control the information flow of the different modalities. \\ \hline
BERT+ViT+Crossmodal Attention (log-Mel spectrogram) \cite{ILIAS2023101485} & Similar to \cite{tsai-etal-2019-multimodal}, the authors exploited a cross-attention mechanism.  \\ \hline
BERT+ViT+Co-Attention \cite{10.3389/fnagi.2022.830943} & The authors used a co-attention mechanism to fuse the representation matrices of the two modalities. \\ \hline
Multimodal BERT - eGeMAPS \cite{10.3389/fnagi.2022.830943} &  The authors injected acoustic information (eGeMAPS) into a BERT model. \\ \hline
Multimodal BERT - ViT \cite{10.3389/fnagi.2022.830943} &  The authors injected image information (via ViT) into a BERT model. \\ \hline
Multimodal BERT - eGeMAPS+ViT \cite{10.3389/fnagi.2022.830943} &  The authors injected both acoustic information (eGeMAPS) and image information (via ViT) into a BERT model. \\ \hline
BERT+ViT+Gated Self-Attention \cite{10.3389/fnagi.2022.830943} & The authors concatenated the outputs of BERT and ViT and passed the resulting matrix through a self-attention layer incorporating a gate model for capturing the inter- and intra-modal interactions. \\ \hline
Transcript+Image+Acoustic \cite{9926818}  & The authors used a Tensor Fusion Layer for fusing the different modalities.  \\ \hline 
\multicolumn{2}{>{\columncolor[gray]{.8}}l}{\textbf{Introduced Approaches without Label Smoothing}} \\
 & Our proposed approaches described in Section \ref{predictive_models} without label smoothing. \\
\bottomrule
\end{tabularx}
\label{baselines_adress_challenge}
\end{table*}

\begin{table*}
\caption{Baselines (ADReSSo Challenge Dataset).}
\begin{tabularx}{\textwidth}{>{\hsize=.32\hsize}X>{\hsize=.68\hsize}X}
\toprule
\multicolumn{1}{c}{\textbf{Reference/Architecture}} & \multicolumn{1}{c}{\textbf{Features/Methodology}} \\ \midrule
\multicolumn{2}{>{\columncolor[gray]{.8}}l}{\textbf{Baselines - Unimodal state-of-the-art approaches (only transcripts)}} \\
BERT & We exploit a BERT model, get the [CLS] token, and pass it through two dense layers consisting of 128 and 2 units respectively. \\ \hline
Model C: LR[Comp]+LR[DisFl]+Ernie+Bert (stacking) \cite{qiao21_interspeech} &  The authors employed model stacking to combine two logistic regression models (LR) using complexity and (dis)fluency features respectively, and the two pretrained language models, i.e. BERT and ERNIE.\\ \hline
Model 5: \cite{pan21c_interspeech} & The authors concatenate the last three states of the BERT sequence classifier with the confidence score input. The confidence score input is generated by the ASR system. \\ \hline
Label Fusion selected models \cite{syed21_interspeech} & The authors extracted a set of handcrafted features, namely syntactic, readability, and lexical diversity, and a set of deep textual embeddings, including BERT and so on. Finally, the authors trained Logistic Regression and SVM classifiers.  \\ \hline
Mp1 \cite{zhu21e_interspeech} & The authors add sentence-level pauses to ASR transcripts and exploit a BERT model.  \\
\midrule
\multicolumn{2}{>{\columncolor[gray]{.8}}l}{\textbf{Baselines - Multimodal state-of-the-art approaches}} \\
LSTM w/ Gating (Words+Acoustic+Disf+Pse+WP) \cite{rohanian21_interspeech} & extraction of acoustic and language features, feed-forward highway layers with gating units  \\ \hline
Global Fusion \cite{pappagari21_interspeech} & fusion of BERT (ASR) and acoustic models, namely x-vectors, x-vectors with 250ms frame-length, and encoder-decoder ASR embeddings (SB Enc/Dec). \\ \hline
Top-10 Avg. \cite{chen21r_interspeech} & Average fusion of predicted class probabilities of the 10 best performing models\\ \hline
Attempt 1: \cite{wang21ca_interspeech} & The authors used acoustic features, linguistic features, and embedding features. For each type of feature, they exploited a deep neural network consisting of multihead attention layers, convolutional layers, and dilated convolutional layers. They used an attention layer for fusing the outputs of the different branches.\\  \midrule
\multicolumn{2}{>{\columncolor[gray]{.8}}l}{\textbf{Introduced Approaches without Label Smoothing}} \\
 & Our proposed approaches described in Section \ref{predictive_models} without label smoothing. \\
\bottomrule
\end{tabularx}
\label{baselines_adressO_challenge}
\end{table*}

\subsection{Experimental Setup}
We divide the ADReSS Challenge train set into a train and a validation set (65\%-35\%). We use a batch size of 4. We train the introduced architectures five times and report the results on the ADReSS Challenge test set via mean $\pm$ standard deviation. Similarly, we divide the ADReSSo Challenge train set into a train and a validation set (65\%-35\%). We train the introduced architectures five times and report the results on the ADReSSo Challenge test set via mean $\pm$ standard deviation. We use \textit{EarlyStopping}, where we stop training if the validation loss has stopped decreasing for eight consecutive epochs. Also, we apply \textit{StepLR} with a step\_size of 4 and a gamma of 0.1. We set $\alpha$ of Eq.~\ref{label_smoothing} equal to 0.001. We set $D=D_c=768$. We set $D_k = D_q = 64$. Regarding the global context strategy, we use one layer of the contextual self-attention mechanism. In terms of the deep-context strategy, we use three layers of the contextual self-attention mechanism. With regards to the deep-global context strategy, we use two layers of the contextual self-attention mechanism. We use the BERT base uncased version and the DeiT\footnote{facebook/deit-base-distilled-patch16-224} model from the Transformers library \cite{wolf-etal-2020-transformers}. For the optimal transport methods, we use the Python library Optimal Transport \cite{flamary2021pot}. All the models have been created using the PyTorch library \cite{NEURIPS2019_9015}. All experiments are conducted on a single Tesla P100-PCIE-16GB GPU.

\subsection{Evaluation Metrics}
\subsubsection{Performance Metrics}

Accuracy (Acc.), Precision (Prec.), Recall (Rec.), F1-Score, and Specificity (Spec.) have been used for evaluating the results of the introduced architectures. These metrics have been computed by regarding the dementia class as the positive one. We report the average and standard deviation of these metrics over five runs.

\subsubsection{Calibration Metrics}

We evaluate the calibration of our model using the metrics proposed by \cite{naeini2015obtaining,Nixon_2019_CVPR_Workshops,pmlr-v70-guo17a}. Specifically, we use the metrics mentioned below:

\begin{itemize}
    \item \textbf{Expected Calibration Error (ECE).} The calibration error is the difference between the fraction of predictions in the bin that are correct (accuracy) and the mean of the probabilities in the bin (confidence). First, we divide the predictions into $M$ equally spaced bins (size $1/M$).

    \begin{equation}
        acc(B_m) = \frac{1}{|B_m|}\sum_{i \in B_m} 1(\hat{y}_i = y_i)
    \end{equation}
    
    \begin{equation}
        conf(B_m) = \frac{1}{|B_m|}\sum_{i \in B_m} \hat{p}_i
    \end{equation}
    , where $y_i$ and $\hat{y}_i$ are the true and predicted labels for the sample $i$ and $\hat{p}_i$ is the confidence (predicted probability value) for sample $i$.
    
    \begin{equation}
        ECE = \sum_{m=1}^{M} \frac{|B_m|}{N} \left|acc\left(B_m \right)-conf\left(B_m \right)\right|
    \end{equation}
    , where $N$ is the total number of data points and $B_m$ is the group of samples whose predicted probability values falls into the interval $I_m=\left(\frac{m-1}{M},\frac{m}{M}\right]$.

    Perfectly calibrated models have an ECE of 0.
    
    \item \textbf{Adaptive Calibration Error (ACE).} Adaptive Calibration Error uses an adaptive scheme which spaces the bin intervals so that each contains an equal number of predictions.

    \begin{equation}
        ACE=\frac{1}{KR} \sum_{k=1}^{K} \sum_{r=1}^{R} |acc(r,k)-conf(r,k)|
    \end{equation}
    , where $acc(r, k)$ and $conf(r, k)$ are the accuracy and confidence of adaptive calibration range $r$ for class label $k$, respectively; and $N$ is the total number of data points. Calibration range $r$ defined by the [$N/R$]th index of the sorted and thresholded predictions.
\end{itemize}

\section{Results}

The results of our introduced models are reported in Tables \ref{Comparison_With_State_of_the_Art_Approaches} and \ref{Comparison_With_State_of_the_Art_Approaches_adresso}. Specifically, Table~\ref{Comparison_With_State_of_the_Art_Approaches} reports the results on the ADReSS Challenge dataset, while Table~\ref{Comparison_With_State_of_the_Art_Approaches_adresso} reports the results on the ADReSSo Challenge dataset. Also, these tables present a comparison of our introduced approaches with existing research initiatives, which have proposed either unimodal or multimodal approaches. In order to compare models, we use the Almost Stochastic Order (ASO) test \cite{del2018optimal, dror2019deep} of statistical significance implemented by \cite{ulmer2022deep}. We use $confidence\_level=0.95$ and $num\_comparisons=50$. Generally, the ASO test determines whether a stochastic order \cite{reimers2018comparing} exists between two models or algorithms, i.e., $A$ and $B$. This method
computes a score $(\epsilon_{min})$ which represents how far the first is from being significantly better in respect
to the second. When $\epsilon_{min}=0$, then one can claim that $A$ is truly stochastically dominant over $B$. When $\epsilon_{min}<0.5$, one can claim that $A$ is almost stochastically dominant over $B$. For $\epsilon_{min}=0.5$, no order can be determined. $(\dagger)$ means that Attention-based Fusion (Deep Context) with label smoothing is stochastically dominant over the respective models. Similarly, in terms of the ADReSSo Challenge dataset, $(\dagger)$ means that Co-Attention (Deep Context) with label smoothing is stochastically dominant over the respective models. $(\star)$ denotes almost stochastic dominance of the Attention-based Fusion (Deep Context) with label smoothing over the respective approaches. Similarly, in terms of the ADReSSo Challenge dataset, $(\star)$ means that Co-Attention (Deep Context) with label smoothing is almost stochastically dominant over the respective models. Note that we cannot compare our approaches with all the existing research initiatives, since we do not have access to the multiple runs or the other approaches have not used multiple runs. In terms of the ECE and ACE calibration metrics, we use ASO for comparing our best performing model, namely Attention-based Fusion (Deep Context) or Co-Attention (Deep Context) with label smoothing, with the respective model without label smoothing.

\subsection{ADReSS Challenge Dataset}

\begin{table*}[!hbt]
\scriptsize
\centering
\caption{Performance comparison among proposed models and state-of-the-art approaches on the ADReSS Challenge test set. Reported values are mean $\pm$ standard deviation. Results are averaged across five runs. $(\dagger)$ means that Attention-based Fusion (Deep Context) with label smoothing is stochastically dominant over the respective models. $(\star)$ denotes almost stochastic dominance of the Attention-based Fusion (Deep Context) with label smoothing over the respective approaches.}
\begin{tabular}{lccccc||cc}
\toprule
\multicolumn{1}{l}{\textbf{Architecture}}&\textbf{Prec. (\%)}&\textbf{Rec. (\%)}&\textbf{F1-score (\%)}&\textbf{Acc. (\%)}&\textbf{Spec. (\%)}&\textbf{ECE}&\textbf{ACE} \\
\midrule
\multicolumn{8}{>{\columncolor[gray]{.8}}l}{\textbf{Baselines - Unimodal state-of-the-art approaches (only transcripts)}} \\
\textit{BERT \cite{9769980}} & 87.19 & 81.66 & 86.73$\dagger$  & 87.50$\dagger$  & 93.33\\
& $\pm$3.25 & $\pm$5.00 & $\pm$4.53 & $\pm$4.37 & $\pm$5.65 \\
\midrule
\multicolumn{8}{>{\columncolor[gray]{.8}}l}{\textbf{ Baselines - Multimodal state-of-the-art approaches}} \\
\textit{\makecell[l]{Fusion Maj. (3-best) \cite{cummins2020comparison}}} &  - & - & 85.40 & 85.20 & - &&\\
\hline
\textit{\makecell[l]{System 3: Phonemes and Audio \cite{edwards20_interspeech}}} & 81.82 & 75.00 & 78.26 & 79.17 & 83.33 &&\\
\hline
\textit{Fusion of system \cite{pompili20_interspeech}} &  94.12 & 66.67 & 78.05 & 81.25 & 95.83 && \\
\hline
\textit{Bimodal Network (Ensembled Output) \cite{koo20_interspeech}} & 89.47 & 70.83 & 79.07 & 81.25 &  91.67 &&\\
\hline
\textit{\makecell[l]{GFI, NUW, Duration, Character 4-grams,\\ Suffixes, POS
tag, UD \cite{martinc20_interspeech}}} & - & - & - &77.08 & - &&\\
\hline
\textit{\makecell[l]{Acoustic \& Transcript \cite{pappagari20_interspeech}}} & 70.00 & 88.00 & 78.00 & 75.00 &  63.00 &&\\
\hline
\textit{Dual BERT \cite{10.3389/fcomp.2021.624683}} & 83.04 & 83.33 & 82.92 & 82.92 &  82.50 &&\\
&  $\pm$3.97 &  $\pm$5.89 &  $\pm$1.86 &  $\pm$1.56 &  $\pm$5.53 &&\\
\hline
\textit{Model C \cite{10.3389/fnagi.2021.623607}} & 78.94 & 62.50 & 69.76 & 72.92 & 83.33 &&\\
 \hline
\textit{\makecell[l]{Majority vote (NLP+Acoustic) \cite{10.3389/fcomp.2021.624659}}} & - & - & - & 83.00 & - &&\\
\hline
 \textit{Audio + Text \cite{9459113}} & - & 87.50 & - & 89.58 & 91.67 && \\
\hline
\textit{\makecell[l]{LSTM with Gating (Acoustic + Lexical + Dis) \cite{rohanian20_interspeech}}} & 81.82 & 75.00 & 78.26 & 79.17 & 83.33 &&\\ \hline
\textit{Ensemble \cite{sarawgi20_interspeech}} & 83.00 & 83.00 & 83.00 & 83.00 & - &&\\ \hline
\textit{{BERT+ViT} \cite{ILIAS2023101485}} & 90.73 & 80.83 & 85.47$\dagger$ & 86.25$\dagger$ & 91.67 \\
\textit{(log-Mel spectrogram)} & $\pm$2.74 & $\pm$2.04 & $\pm$1.70 & $\pm$1.67 & $\pm$2.64 \\ \hline
\textit{{BERT+ViT+Gated Multimodal Unit} \cite{ILIAS2023101485}} & 80.92 & 91.67 & 85.92$\dagger$ & 85.00$\dagger$ & 78.33 \\
\textit{(log-Mel spectrogram)} & $\pm$2.30 & $\pm$3.73 & $\pm$2.37 & $\pm$2.43 & $\pm$3.12 \\ \hline
\textit{{BERT+ViT+Crossmodal Attention} \cite{ILIAS2023101485}} & 86.13 & 91.67 & 88.69$\star$ & 88.33$\star$ & 85.00 \\
\textit{(log-Mel spectrogram)} & $\pm$3.26 & $\pm$4.56 & $\pm$2.12 & $\pm$2.12 & $\pm$4.25 \\ \hline
 \textit{{BERT+ViT+Co-Attention} \cite{10.3389/fnagi.2022.830943}} & 92.83 & 81.67 & 86.81$\star$ & 87.50$\star$ & 93.33 &&\\
&  $\pm$6.39 &  $\pm$2.04 &  $\pm$3.37 &  $\pm$3.49 &  $\pm$6.24 &&\\ \hline
 \textit{{Multimodal BERT - eGeMAPS} \cite{10.3389/fnagi.2022.830943}} & 74.51 & 87.50 & 80.35$\dagger$ & 78.75$\dagger$ & 70.00 && \\ 
&  $\pm$1.01 &  $\pm$6.45 &  $\pm$2.77 &  $\pm$2.04 &  $\pm$3.12 && \\ \hline
 \textit{{Multimodal BERT - ViT} \cite{10.3389/fnagi.2022.830943}} & 73.91 & 91.67 & 81.79$\dagger$ & 79.58$\dagger$ & 67.50 && \\ 
&  $\pm$2.40 &  $\pm$2.64 &  $\pm$1.72 &  $\pm$2.04 &  $\pm$4.08 && \\ \hline
 \textit{{Multimodal BERT - eGeMAPS+ViT} \cite{10.3389/fnagi.2022.830943}} & 76.57 & 89.17 & 82.28$\dagger$ & 80.83$\dagger$ & 72.50 && \\ 
&  $\pm$3.74 &  $\pm$5.65 &  $\pm$3.49 &  $\pm$3.58 &  $\pm$5.65 && \\ \hline
 \textit{{BERT+ViT+Gated Self-Attention} \cite{10.3389/fnagi.2022.830943}} & 90.87 & 89.17 &  89.94$\star$ &  90.00$\star$ & 90.83 && \\
&  $\pm$3.50 &  $\pm$2.04 &  $\pm$1.36 &  $\pm$1.56 &  $\pm$4.08 && \\ \hline
\textit{Transcript+Image+Acoustic \cite{9926818}} & 90.88 & 80.83 & 85.48$\dagger$ & 86.25$\dagger$ & 91.66 \\
& $\pm$3.60 &  $\pm$2.04 &  $\pm$0.76 &  $\pm$1.02 &  $\pm$3.73 \\ 

\midrule
\multicolumn{8}{>{\columncolor[gray]{.8}}l}{\textbf{ Baselines - Introduced models (without label smoothing)}} \\
\textit{{Co-Attention}} &  89.62 &  85.83 &  87.63$\dagger$ & 87.92$\dagger$ &  90.00 & 0.1208 & 0.1660\\ 
\textit{(Global Context)} &  $\pm$1.75 &  $\pm$3.33 &  $\pm$1.80 &  $\pm$1.56 &  $\pm$2.04 & $\pm$0.2296 & $\pm$0.0335\\ \hline
\textit{{Co-Attention}} &  88.25 & 87.50 & 87.85$\star$ & 87.92$\dagger$ & 88.33 & 0.1384 & 0.1532\\ 
\textit{(Deep Context)} &  $\pm$1.56 &  $\pm$2.64 &  $\pm$1.66 &  $\pm$1.56 &  $\pm$1.66 & $\pm$0.0109 & $\pm$0.0110\\ \hline
\textit{{Co-Attention}} & 90.26 & 85.00 & 87.51$\star$ & 87.92$\star$ & 90.83 & 0.1355 & 0.1648\\ 
\textit{(Deep-Global Context)} &  $\pm$1.70 &  $\pm$4.25 &  $\pm$2.69 &  $\pm$2.43 &  $\pm$1.66 & $\pm$0.0183 & $\pm$0.0119\\ \hline
\textit{{Attention-based Fusion}} & 89.55 & 85.83 & 87.32$\star$ & 87.50$\star$ & 89.16 & 0.1256 & 0.1279\\ 
 \textit{(Global Context)} &  $\pm$7.31 &  $\pm$6.24 &  $\pm$4.35 &  $\pm$4.37 &  $\pm$8.58 & $\pm$0.0291 & $\pm$0.0277 \\ \hline
 \textit{{Attention-based Fusion}} & 91.06 & 89.16 & 89.95$\star$ & 90.00$\star$ & 90.83 & 0.0975$\star$ & 0.1046$\star$ \\ 
 \textit{(Deep Context)} &  $\pm$5.04 &  $\pm$3.33 &  $\pm$1.91 &  $\pm$2.04 &  $\pm$5.53 & $\pm$0.0188 & $\pm$0.0173\\ \hline
  \textit{{Attention-based Fusion}} & 90.45 & 85.83 & 88.04$\star$ & 88.33$\star$ & 90.83 & 0.1173 & 0.1065 \\ 
 \textit{(Deep-Global Context)} &  $\pm$2.93 &  $\pm$2.04 &  $\pm$1.65 &  $\pm$1.66 &  $\pm$3.12 & $\pm$0.0134 & $\pm$0.0153 \\
 \midrule
\multicolumn{8}{>{\columncolor[gray]{.8}}l}{\textbf{ Introduced models (with label smoothing)}} \\
 \textit{{Co-Attention}} & 88.65 & 88.33 & 88.39$\star$ & 88.33$\star$ & 88.33 & 0.1075 & 0.1710\\ 
\textit{(Global Context)} &  $\pm$4.63 &  $\pm$1.66 &  $\pm$1.76 &  $\pm$2.12 &  $\pm$5.53 & $\pm$0.0198 & $\pm$0.0281 \\ \hline
\textit{{Co-Attention}} &  93.57 & 84.16 & 88.53$\star$ & 89.16$\star$ & 94.16 & 0.1082 & 0.1316\\ 
\textit{(Deep Context)} &  $\pm$2.08 &  $\pm$4.86 &  $\pm$2.79 &  $\pm$2.43 &  $\pm$2.04 & $\pm$0.0184 & $\pm$0.0296 \\ \hline
\textit{{Co-Attention}} &  87.88 &  87.50 &  87.39$\star$ & 87.50$\dagger$ &  87.50 & 0.1176 & 0.1568 \\ 
\textit{(Deep-Global Context)} &  $\pm$3.73 &  $\pm$6.97 &  $\pm$2.45 &  $\pm$1.86 &  $\pm$4.56 & $\pm$0.0167 & $\pm$0.0306\\ \hline
\textit{{Attention-based Fusion}} &  90.51 &  85.00 & 87.53$\dagger$ & 87.92$\dagger$ & 90.83 & 0.1094 & 0.1168\\ 
 \textit{(Global Context)} &  $\pm$3.40 &  $\pm$4.25 &  $\pm$1.75 &  $\pm$1.56 &  $\pm$4.08 & $\pm$0.0086 & $\pm$0.0099\\ \hline
 \textit{{Attention-based Fusion}} & 93.08 & 89.17 & 91.06 & 91.25 & 93.33 & 0.0859 & 0.0830 \\ 
 \textit{(Deep Context)} &  $\pm$2.03 &  $\pm$2.04 &  $\pm$1.60 &  $\pm$1.56 &  $\pm$2.04 & $\pm$0.0130 & $\pm$0.0158 \\ \hline
  \textit{{Attention-based Fusion}} & 89.87 & 83.33 & 86.20$\dagger$ & 86.66$\dagger$ & 90.00 & 0.1397 & 0.1508\\ 
 \textit{(Deep-Global Context)} &  $\pm$5.52 &  $\pm$4.56 &  $\pm$0.90 &  $\pm$1.02 &  $\pm$5.65 & $\pm$0.0102 & $\pm$0.0123 \\ 
 \bottomrule
\end{tabular}
%\label{compare}
\label{Comparison_With_State_of_the_Art_Approaches}
\end{table*}

Regarding our proposed models, one can observe that Attention-based Fusion (Deep Context) constitutes our best performing model outperforming all the other introduced models in all the evaluation metrics except Precision and Specificity. Specifically, Attention-based Fusion (Deep Context) outperforms the other introduced models with label smoothing in Accuracy by 2.09-4.59\%, in Recall by 0.84-5.84\%, and in F1-score by 2.53-4.86\%. Despite the fact that Attention-based Fusion (Deep Context) obtains a lower Precision score by other introduced models, it surpasses them in F1-score, which constitutes the weighted average of Precision and Recall. Although it achieves lower specificity scores by Co-Attention (Deep Context), it must be noted that in health related studies, F1-score is more important than Specificity, since high F1-score means that the model can detect better the AD patients, while high Specificity and low F1-score means that AD patients are misdiagnosed as non-AD ones. In addition, Co-Attention (Deep Context) constitutes our second best performing model attaining an Accuracy of 89.16\%. It achieves the highest precision and specificity scores accounting for 93.57\% and 94.16\% respectively, while it achieves an F1-score of 88.53\%. It outperforms all the introduced models, except Attention-based Fusion (Deep Context), in Accuracy by 0.83-2.50\% and in F1-score by 0.14-2.33\%. It outperforms all the models in Precision and Specificity by 0.49-5.69\% and 0.83-6.66\% respectively. 

Next, we compare our introduced approaches with label smoothing with the ones without applying label smoothing. As one can easily observe, label smoothing leads to both performance improvement and better calibration of the proposed approaches. Specifically, we observe that Attention-based Fusion (Deep Context) with label smoothing obtains a higher Accuracy score than the one obtained by the respective model without label smoothing by 1.25\%, Attention-based Fusion (Global Context) with label smoothing surpasses Attention-based Fusion (Global Context) without label smoothing in Accuracy by 0.42\%, etc. In terms of the calibration metrics, namely ECE and ACE, one can observe that label smoothing leads to better calibrated models. For instance, Attention-based Fusion (Deep Context) with label smoothing obtains an ECE of 0.0859 and an ACE of 0.0830, which are significantly better than the ones obtained by Attention-based Fusion (Deep Context) without label smoothing by 0.0116 and 0.0216 respectively.

In comparison with the unimodal and multimodal state-of-the-art approaches, one can observe that our best performing model, namely Attention-based Fusion (Deep Context) with label smoothing, outperforms the research works in Accuracy by 1.25-18.33\% and in F1-score by 1.12-21.30\%. These differences in performance are attributable to the fact that our best performing model captures both the inter- and intra-modal interactions through the self-attention mechanisms and optimal transport domain adaptation methods, enhances the self-attention mechanism with contextual information, and applies label smoothing in contrast to the research initiatives. In addition, Co-Attention (Deep Context) outperforms the research works, except \cite{10.3389/fnagi.2022.830943,9459113}, in Accuracy by 0.83-16.24\%. 

\subsection{ADReSSo Challenge Dataset}

As one can easily observe in Table~\ref{Comparison_With_State_of_the_Art_Approaches_adresso}, Co-Attention (Deep Context) with label smoothing constitutes our best performing model attaining an Accuracy of 85.35\%, a Recall of 86.29\%, and a F1-score of 85.27\%. It surpasses the other introduced models (with label smoothing) in Accuracy by 1.41-4.22\%, in Recall by 1.15-4.58\%, and in F1-score by 1.92-4.32\%. In addition, we observe that Co-Attention (Deep Context) with label smoothing achieves better performance than the one obtained by the respective model without label smoothing. Specifically, the Accuracy is improved by 2.25\%, the Recall is improved by 2.29\%, the F1-score presents a surge of 2.26\%, the Precision is increased by 2.21\%, and the Specificity is improved by 2.21\%. In terms of the calibration metrics, we observe that the ECE is improved by 0.0171 (ASO test indicates almost stochastic dominance). 

Comparing our introduced models with label smoothing with the ones without label smoothing, we observe that in most cases label smoothing contributes to both the performance improvement and better calibration. For instance, Co-Attention (Global Context) with label smoothing improves Accuracy by 1.12\% compared with the respective model without label smoothing, while ECE and ACE are also improved by 0.0254 and 0.0387 respectively. Similarly, Attention-based Fusion (Deep Context) with label smoothing outperforms the respective model without label smoothing in F1-score and Accuracy by 2.27\% and 2.53\% respectively, while the ECE and ACE also present a decline of 0.0106 and 0.0077 respectively. 

In comparison with the unimodal and multimodal baselines, we observe that our best performing model, namely Co-Attention (Deep Context) with label smoothing, outperforms these baselines in Accuracy by 0.84-5.35\%. Also, it outperforms all the research works, except for \cite{chen21r_interspeech}, in F1-score by 0.34-6.74\%. We observe that our best performing model attains a better performance than BERT (ASO test indicates stochastic dominance), verifying our initial hypothesis that both modalities, i.e., transcripts and audio files, contribute to a better performance. In addition, we observe that our second best performing model, namely Attention-based Fusion (Deep Context) outperforms some research works, except for \cite{pan21c_interspeech,syed21_interspeech,rohanian21_interspeech, pappagari21_interspeech}, in Accuracy by 0.84-3.94\%.

\begin{table*}[!hbt]
\scriptsize
\centering
\caption{Performance comparison among proposed models and state-of-the-art approaches on the ADReSSo Challenge test set. Reported values are mean $\pm$ standard deviation. Results are averaged across five runs. $(\dagger)$ means that Co-Attention (Deep Context) with label smoothing is stochastically dominant over the respective models. $(\star)$ denotes almost stochastic dominance of the Co-Attention (Deep Context) with label smoothing over the respective approaches.}
\begin{tabular}{lccccc||cc}
\toprule
\multicolumn{1}{l}{\textbf{Architecture}}&\textbf{Prec. (\%)}&\textbf{Rec. (\%)}&\textbf{F1-score (\%)}&\textbf{Acc. (\%)}&\textbf{Spec. (\%)}&\textbf{ECE}&\textbf{ACE} \\
\midrule
\multicolumn{8}{>{\columncolor[gray]{.8}}l}{\textbf{Baselines - Unimodal state-of-the-art approaches (only transcripts)}} \\
 \textit{BERT} & 83.35 & 74.29 & 78.53$\dagger$ & 80.00$\dagger$ & 85.55 & - & - \\
&  $\pm$0.86 &  $\pm$2.55 &  $\pm$1.43 &  $\pm$1.05 &  $\pm$1.11 & -& -\\ \hline
\textit{Model C:} \cite{qiao21_interspeech} & 85.00 & 80.00 & 82.00 & 83.00 & 86.00 & - & - \\ \hline
\textit{Model 5:} \cite{pan21c_interspeech} & 81.58 & 88.57 & 84.93 & 84.51 & 80.56 & - & - \\ \hline
\textit{Label Fusion selected models} \cite{syed21_interspeech} & - & - & - & 84.51 & - & - & - \\ \hline
\textit{Mp1} \cite{zhu21e_interspeech} & 87.10 & 77.14 & 81.82 & 83.10 & 88.89 & - & - \\
\midrule
\multicolumn{8}{>{\columncolor[gray]{.8}}l}{\textbf{ Baselines - Multimodal state-of-the-art approaches}} \\ 
\textit{LSTM w/ Gating (Words+Acoustic+Disf+Pse+WP)} \cite{rohanian21_interspeech} & - & - & - & 84.00 & - & - & - \\ \hline
\textit{Global Fusion} \cite{pappagari21_interspeech} & 92.00 & 74.00 & 83.00 & 84.51 & 94.00 & - & - \\ \hline
\textit{Top-10 Avg.} \cite{chen21r_interspeech} & - & - & 88.89 & 81.69 & 80.00 & - & - \\ \hline
\textit{Attempt 1:} \cite{wang21ca_interspeech} & 75.00 & 91.67 & 82.50 & 80.28 & 68.57 & - & - \\

\midrule
\multicolumn{8}{>{\columncolor[gray]{.8}}l}{\textbf{ Baselines - Introduced models (without label smoothing)}} \\
 \textit{{Co-Attention}} & 83.77 & 81.13 & 81.85$\star$ & 82.54$\star$ & 83.88 & 0.1536 & 0.2017\\ 
 \textit{(Global Context)} &  $\pm$4.59 &  $\pm$9.13 &  $\pm$3.01 &  $\pm$1.69 &  $\pm$6.66 & $\pm$0.0311 & $\pm$0.0214 \\ \hline
 \textit{{Co-Attention}} & 82.22  &  84.00 &  83.01$\star$ & 83.10$\star$& 82.22  & 0.1349$\star$ & 0.1845\\ 
 \textit{(Deep Context)} &  $\pm$1.79 &  $\pm$4.28 &  $\pm$1.63 &  $\pm$1.26 &  $\pm$2.83 & $\pm$0.0135 & $\pm$0.0169\\ \hline
\textit{{Co-Attention}} & 83.03  & 80.57  & 81.73$\star$  & 82.25$\dagger$& 83.88  & 0.1414 & 0.1948\\ 
 \textit{(Deep-Global Context)} &  $\pm$2.07 &  $\pm$2.79 &  $\pm$1.23 &  $\pm$1.13 &  $\pm$2.72 & $\pm$0.0091 & $\pm$0.0265\\ \hline
 \textit{{Attention-based Fusion}} &  83.44 &  74.86 &  78.90$\dagger$ & 80.28$\dagger$ & 85.56  & 0.1633 & 0.1825\\ 
 \textit{(Global Context)} &  $\pm$1.16 &  $\pm$2.14 &  $\pm$1.51 &  $\pm$1.26 &  $\pm$1.11 & $\pm$0.0207 & $\pm$0.0140\\ \hline
 \textit{{Attention-based Fusion}} & 81.52  & 81.14  & 81.08$\dagger$  & 81.41$\dagger$&  81.66 & 0.1442 &0.1737 \\ 
 \textit{(Deep Context)} &  $\pm$3.47 &  $\pm$5.59 &  $\pm$1.58 &  $\pm$1.05 &  $\pm$5.44 & $\pm$0.0284 & $\pm$0.0089\\ \hline
 \textit{{Attention-based Fusion}} & 79.58  & 85.71  &  82.38$\star$ & 81.97$\dagger$&  78.33 & 0.1671 & 0.1820\\ 
\textit{(Deep-Global Context)} &  $\pm$2.69 &  $\pm$4.78 &  $\pm$1.59 &  $\pm$1.38 &  $\pm$4.78 & $\pm$0.0201 & $\pm$0.0193\\ 
 \midrule
\multicolumn{8}{>{\columncolor[gray]{.8}}l}{\textbf{ Introduced models (with label smoothing)}} \\
 \textit{{Co-Attention}} & 84.77  & 81.71 &  83.12$\star$ & 83.66$\star$ &  85.55 & 0.1282 & 0.1630\\ 
 \textit{(Global Context)} &  $\pm$2.39 &  $\pm$3.43 &  $\pm$0.95 &  $\pm$0.69 &  $\pm$3.24 & $\pm$0.0053 & $\pm$0.0179\\ \hline
\textit{{Co-Attention}} & 84.43 & 86.29  & 85.27 & 85.35 & 84.43 & 0.1178 & 0.1800\\ 
\textit{(Deep Context)} &  $\pm$1.59 &  $\pm$4.19 &  $\pm$1.78 &  $\pm$1.44 &  $\pm$2.19 & $\pm$0.0209 & $\pm$0.0213 \\ \hline
 \textit{{Co-Attention}} & 82.45  &  82.86 &  82.55$\star$ & 82.82$\star$ &  82.77 & 0.1443 & 0.1749\\ 
\textit{(Deep-Global Context)} &  $\pm$0.99 &  $\pm$4.78 &  $\pm$2.03 &  $\pm$1.38 &  $\pm$2.08 & $\pm$0.0046 & $\pm$0.0082\\ \hline
\textit{{Attention-based Fusion}} &  80.44 & 81.71  &  80.95$\dagger$ & 81.13$\dagger$&  80.55 & 0.1540 & 0.1920\\ 
\textit{(Global Context)} &  $\pm$1.65 &  $\pm$4.98 &  $\pm$2.04 &  $\pm$1.44 &  $\pm$3.04 & $\pm$0.0195 & $\pm$0.0215\\ \hline
\textit{{Attention-based Fusion}} &  85.10 &  81.71 &  83.35$\star$ & 83.94$\star$& 86.11  & 0.1336 & 0.1660\\ 
\textit{(Deep Context)} &  $\pm$0.53 &  $\pm$3.43 &  $\pm$2.04 &  $\pm$1.69 &  $\pm$0.04 & $\pm$0.0190 & $\pm$0.0144\\ \hline
\textit{{Attention-based Fusion}} &  81.45 & 85.14  & 83.18$\star$  & 83.10$\star$&  81.11 & 0.1690 & 0.1938\\ 
\textit{(Deep-Global Context)} &  $\pm$1.32 &  $\pm$4.92 &  $\pm$2.42 &  $\pm$1.99 &  $\pm$2.08 & $\pm$0.0245 & $\pm$0.0112\\ 
 \bottomrule
\end{tabular}
%\label{compare}
\label{Comparison_With_State_of_the_Art_Approaches_adresso}
\end{table*}

\section{Ablation Study}

In this section, we run a series of ablation experiments using the ADReSS Challenge dataset to explore the effectiveness and robustness of the introduced architecture described in Section \ref{predictive_models}. We report the results of the ablation study in Tables~\ref{Ablation_Study_1} and \ref{Ablation_Study_2}.

First, we explore the effectiveness of the context-based self-attention. To do this, we remove the contextual information and exploit the conventional self-attention mechanism introduced by \cite{10.5555/3295222.3295349}. We observe that the Accuracy score drops from 91.25\% to 87.08\%, while the F1-score presents a decline of 4.60\%. Also, we observe that the removal of contextual information yields to higher standard deviations of the performance metrics.

Next, we investigate the efficacy of the gate model, which is incorporated into the self-attention mechanism. To do this, we remove the gate model and exploit the conventional self-attention mechanism. We observe that Accuracy and F1-score present a decline of 2.50\% and 2.84\% respectively.

Moreover, we explore the effectiveness of the optimal transport domain adaptation method and the Optimal Transport Kernel. To do this, we remove these components from the introduced architecture. We observe that the Accuracy score is equal to 87.50\%, which is lower by 3.75\% than the one obtained by our best performing model. Also, this approach yields an F1-score accounting for 87.47\%, which is lower by 3.59\% than the one achieved by Attention-based Fusion (Deep Context).

Next, we explore the effectiveness of the Optimal Transport Kernel. To do this, we remove this component, exploit the average operation over the sequence length, and finally repeat the vector $n$ times, so as to ensure that both the textual and image modalities have the same sequence length. As one can observe, this method presents a decline in Accuracy score by 2.92\%, while the F1-score is also reduced by 2.33\%.

In addition, we explore the effectiveness of the fusion method. To prove this, we remove the fusion method, apply the average operation over $C$ (Eq.~\ref{concat_text}) and $S$ (Eq.~\ref{concat_image}) and concatenate these two representation vectors. We observe that the concatenation of features yields an Accuracy and F1-score of 87.50\% and 87.65\% respectively. This difference in performance can be justified by the fact that the concatenation operation does not capture the inherent correlations between the modalities.

Finally, we vary the layers of the context-based self-attention mechanism. The results of this ablation study are reported in Table~\ref{Ablation_Study_2}. As the number of layers increases from 1 to 3, we observe that both the Accuracy and F1-score also increase. This justifies our initial hypothesis that stacking attention layers and fusing the outputs of different layers into one context vector, yields to better evaluation results, since the model captures both high-level and low-level syntactic and semantic information. However, we observe that the performance of our approach starts to present a decline by stacking four or five layers of context-based self-attention by applying the deep-context strategy. We assume that this decline in performance is attributable to the limited dataset used and consequently to the problem of overfitting.

\begin{table*}[!hbt]
\scriptsize
\centering
\caption{Ablation Study. Reported values are mean $\pm$ standard deviation. Results are averaged across five runs.}
\begin{tabular}{lccccc}
\toprule
\multicolumn{1}{l}{\textbf{ Architecture}}&\textbf{ Prec. (\%)}&\textbf{ Rec. (\%)}&\textbf{ F1-score (\%)}&\textbf{ Acc. (\%)}&\textbf{ Spec. (\%)}\\
\midrule
 \textit{{without contextual vector in self-attention}} & 91.34 & 83.33 & 86.46 & 87.08 & 90.83 \\
&  $\pm$7.35 &  $\pm$9.50 &  $\pm$4.64 &  $\pm$4.04 &  $\pm$10.00 \\ \hline
  \textit{{self-attention without gate model}} &  92.99 &  84.16 &  88.22 & 88.75 &  93.33  \\ 
   &  $\pm$4.28 &  $\pm$3.12 &  $\pm$0.88 &  $\pm$1.02 &  $\pm$4.25 \\ \hline
 \textit{without optimal transport and OTK} &  87.60 & 87.50 & 87.47 & 87.50 & 87.50 \\ 
  &  $\pm$2.02 &  $\pm$3.73 &  $\pm$1.52 &  $\pm$1.32 &  $\pm$2.64 \\ \hline
\textit{repeat vector instead of OTK} & 86.08 & 91.66 & 88.73 & 88.33 & 85.00 \\ 
 &  $\pm$3.37 &  $\pm$2.64 &  $\pm$1.97 &  $\pm$2.12 &  $\pm$4.25 \\ \hline
 \textit{{Concatenation - Without fusion}} &  87.23 & 88.33 & 87.65 & 87.50 & 86.66 \\ 
  &  $\pm$4.99 &  $\pm$3.12 &  $\pm$2.64 &  $\pm$2.95 &  $\pm$6.12 \\ \hline
\textit{{Proposed Framework}} &\textbf{93.08} & \textbf{89.17} & \textbf{91.06} & \textbf{91.25} & \textbf{93.33} \\ 
 & $\pm$\textbf{2.03} &  $\pm$\textbf{2.04} &  $\pm$\textbf{1.60} &  $\pm$\textbf{1.56} &  $\pm$\textbf{2.04} \\ 

\bottomrule
\end{tabular}
%\label{compare}
\label{Ablation_Study_1}
\end{table*}

\begin{table*}[!hbt]
\scriptsize
\centering
\caption{Ablation Study. Reported values are mean $\pm$ standard deviation. Results are averaged across five runs.}
\begin{tabular}{lccccc}
\toprule
\textbf{Layers}&\textbf{ Prec. (\%)}&\textbf{ Rec. (\%)}&\textbf{ F1-score (\%)}&\textbf{ Acc. (\%)}&\textbf{ Spec. (\%)}\\
\midrule
\textit{1} &  90.37 &  83.33 &  86.52 & 87.08 &  90.83 \\ 
 &  $\pm$3.33 &  $\pm$5.27 &  $\pm$1.94 &  $\pm$1.56 &  $\pm$4.08 \\ \hline
\textit{2} & 88.09 & 91.66 & 89.77 & 89.58 & 87.50  \\ 
 &  $\pm$1.96 &  $\pm$3.73 &  $\pm$1.45 &  $\pm$1.32 &  $\pm$2.64 \\ \hline
 \textbf{\textit{3}} & \textbf{93.08} & \textbf{89.17} & \textbf{91.06} & \textbf{91.25} & \textbf{93.33}  \\ 
 \textbf{(Our best performing model)}&  $\pm$\textbf{2.03} &  $\pm$\textbf{2.04} &  $\pm$\textbf{1.60} &  $\pm$\textbf{1.56} &  $\pm$\textbf{2.04} \\ \hline
\textit{4} & 92.05 & 76.66 & 83.55 & 85.00 & 93.33 \\
 &  $\pm$3.70 &  $\pm$5.65 &  $\pm$4.28 &  $\pm$3.58 &  $\pm$3.33 \\ \hline
 \textit{5} & 88.67 & 83.33 & 85.84 & 86.25 & 89.16 \\
 &  $\pm$4.20 &  $\pm$3.73 &  $\pm$2.86 &  $\pm$2.83 &  $\pm$4.25 \\
\bottomrule
\end{tabular}
\label{Ablation_Study_2}
\end{table*}

\section{Discussion}
 
From the results obtained in this study, we found that:
\begin{itemize}
    \item \textbf{Finding 1:} We proposed a context-based self-attention mechanism and exploited three approaches of adding contextual information to self-attention layers. Results on the ADReSS and ADReSSo Challenge datasets showed that the fusion of the outputs (low-level syntactic and semantic information) of different layers as a deep context vector yielded the highest evaluation results.
    \item \textbf{Finding 2:} We compared our proposed approaches with and without label smoothing. Findings suggested that label smoothing contributes to both the performance improvement and improvements in terms of the calibration metrics.
    \item \textbf{Finding 3:} We exploited two methods for fusing the self and cross-attention features. Findings of the experiments conducted on the ADReSS Challenge dataset suggested that the usage of two independent attentional reduction models, the add operation, and the layer normalization achieved better performance than the usage of a co-attention mechanism. On the other hand, results on the ADReSSo Challenge dataset showed that the co-attention mechanism as a fusion method achieved the best evaluation results.
    \item \textbf{Finding 4:} Findings from a series of ablation studies showed the effectiveness and robustness of the introduced architecture.
    \item \textbf{Finding 5:} Our proposed models yielded competitive performances to the existing state-of-the-art approaches. We also used the Almost Stochastic Order test to test for statistical significance. This test does not make any assumptions about the distributions of the scores.
     \item \textbf{Finding 6:} We observed that in most cases the performance of the multi-modal models (baselines) was inferior to the transcript only BERT baseline. We hypothesize that this difference in performance is attributable to the fact that the multimodal approaches propose early and late fusion strategies or add / concatenate the representation vectors of different modalities during training. In this way, the inter-modal interactions cannot be captured effectively. This difference in performance justifies our initial motivation that more effective fusion methods must be explored for capturing the inter-modal interactions.
\end{itemize}

Our approaches entail some limitations, which are described below:

\begin{itemize}
    \item Hyperparameter Tuning: In this study, we did not perform hyperparameter tuning due to the limited access to GPU resources. Hyperparameter tuning yields to an increase of the classification performance.
    \item Explainability: In this study, we did not apply explainability techniques, namely LIME, Integrated Gradients, and so on, for explaining the predictions of our introduced approaches. 
    \item Self-Supervised Learning: Contrary to self-supervised learning methods, our approach relies heavily on plenty of training data.
\end{itemize}

\section{Conclusion and Future Work}

In this paper, we introduced some new approaches to detect AD patients from speech and transcripts, which capture the inter- and intra-modal interactions, enhance the conventional self-attention mechanism with contextual information, and deal with the problem of creating overconfident models by applying label smoothing. Our proposed architectures consist of BERT, DeiT, self-attention mechanism incorporating a gating model, context-based self-attention, optimal transport domain adaptation methods, and one new method for fusing the self and cross-attention features in the task of dementia detection from speech data. Furthermore, we designed extensive ablation experiments to explore the effectiveness of the components of the proposed architecture. Extensive experiments conducted on the ADReSS Challenge dataset demonstrate the efficacy of the proposed architectures reaching Accuracy and F1-score up to 91.25\% and 91.06\% respectively. Also, findings suggested that the label smoothing contributes to both the performance improvement and calibration of our model.

We evaluated our proposed approaches on the ADReSS Challenge dataset, which consists of a statistically balanced and acoustically enhanced set of recordings of spontaneous speech sessions. The proposed models can be embedded into an application, which will be capable of detecting AD patients with high accuracy. The creation of such an application appears to be very important especially in the era of covid-19, where there are restrictions for access to hospitals and medical centers.

In the future, we plan to exploit our methods in other health-related tasks, including the detection of Parkinson's disease. This will establish the generalizability of our introduced architectures. In addition, we aim to use automatic transcripts, since there are datasets, which do not include manual transcripts \cite{luz21_interspeech,luz2023multilingual}. Also, we plan to use the ADReSS-M Challenge dataset \cite{luz2023multilingual}, where we will train our proposed approaches using automatic transcripts and audio files in the english language and test our introduced approach using automatic transcripts and audio files in the greek language. We also plan to employ more methods for measuring the uncertainty and calibration of our models. Applying self-supervised learning methods with limited speech data is one of our plans. In this way, our methods will not rely on labels and training data. Additionally, inspired by \cite{lu2021detection}, we aim to exploit AlexNet instead of DeiT. Specifically, similar to \cite{lu2021detection}, we plan to substitute the last several layers in AlexNet with extreme learning machine and optimize the extreme learning machine with a chaotic bat algorithm. Finally, motivated by \cite{https://doi.org/10.1002/int.22686}, we aim to apply some of the modules to our task, including the image-level representation learning algorithm, the universal neighboring-aware representation learning framework, and the graph random vector functional link.

%\section{}\label{}

\printcredits

\bibliographystyle{cas-model2-names}

\bibliography{cas-refs}

\begin{thebibliography}{92}
\expandafter\ifx\csname natexlab\endcsname\relax\def\natexlab#1{#1}\fi
\providecommand{\url}[1]{\texttt{#1}}
\providecommand{\href}[2]{#2}
\providecommand{\path}[1]{#1}
\providecommand{\DOIprefix}{doi:}
\providecommand{\ArXivprefix}{arXiv:}
\providecommand{\URLprefix}{URL: }
\providecommand{\Pubmedprefix}{pmid:}
\providecommand{\doi}[1]{\href{http://dx.doi.org/#1}{\path{#1}}}
\providecommand{\Pubmed}[1]{\href{pmid:#1}{\path{#1}}}
\providecommand{\bibinfo}[2]{#2}
\ifx\xfnm\relax \def\xfnm[#1]{\unskip,\space#1}\fi
%Type = Inproceedings
\bibitem[{Al-Hameed et~al.(2017)Al-Hameed, Benaissa and
  Christensen}]{10.1145/3175587.3175589}
\bibinfo{author}{Al-Hameed, S.}, \bibinfo{author}{Benaissa, M.},
  \bibinfo{author}{Christensen, H.}, \bibinfo{year}{2017}.
\newblock \bibinfo{title}{Detecting and predicting alzheimer's disease severity
  in longitudinal acoustic data}, in: \bibinfo{booktitle}{Proceedings of the
  International Conference on Bioinformatics Research and Applications 2017},
  \bibinfo{publisher}{Association for Computing Machinery},
  \bibinfo{address}{New York, NY, USA}. p. \bibinfo{pages}{57–61}.
\newblock \URLprefix \url{https://doi.org/10.1145/3175587.3175589},
  \DOIprefix\doi{10.1145/3175587.3175589}.
%Type = Misc
\bibitem[{{Alzheimer's Society}({2021})}]{dementia_language}
\bibinfo{author}{{Alzheimer's Society}}, \bibinfo{year}{{2021}}.
\newblock \bibinfo{title}{{\textit{Dementia and language}}}.
\newblock \bibinfo{howpublished}{Available online at:
  \url{https://www.alzheimers.org.uk/about-dementia/symptoms-and-diagnosis/symptoms/dementia-and-language}}.
%Type = Article
\bibitem[{Arevalo et~al.(2020)Arevalo, Solorio, Montes-y Gomez and
  Gonz{\'a}lez}]{arevalo2020gated}
\bibinfo{author}{Arevalo, J.}, \bibinfo{author}{Solorio, T.},
  \bibinfo{author}{Montes-y Gomez, M.}, \bibinfo{author}{Gonz{\'a}lez, F.A.},
  \bibinfo{year}{2020}.
\newblock \bibinfo{title}{Gated multimodal networks}.
\newblock \bibinfo{journal}{Neural Computing and Applications} ,
  \bibinfo{pages}{1--20}.
%Type = Article
\bibitem[{Ba et~al.(2016)Ba, Kiros and Hinton}]{ba2016layer}
\bibinfo{author}{Ba, J.L.}, \bibinfo{author}{Kiros, J.R.},
  \bibinfo{author}{Hinton, G.E.}, \bibinfo{year}{2016}.
\newblock \bibinfo{title}{Layer normalization}.
\newblock \bibinfo{journal}{arXiv preprint arXiv:1607.06450} .
%Type = Article
\bibitem[{Becker et~al.(1994)Becker, Boiler, Lopez, Saxton and
  McGonigle}]{10.1001/archneur.1994.00540180063015}
\bibinfo{author}{Becker, J.T.}, \bibinfo{author}{Boiler, F.},
  \bibinfo{author}{Lopez, O.L.}, \bibinfo{author}{Saxton, J.},
  \bibinfo{author}{McGonigle, K.L.}, \bibinfo{year}{1994}.
\newblock \bibinfo{title}{{The Natural History of Alzheimer's Disease:
  Description of Study Cohort and Accuracy of Diagnosis}}.
\newblock \bibinfo{journal}{Archives of Neurology} \bibinfo{volume}{51},
  \bibinfo{pages}{585--594}.
\newblock \URLprefix
  \url{https://doi.org/10.1001/archneur.1994.00540180063015},
  \DOIprefix\doi{10.1001/archneur.1994.00540180063015},
  \href{http://arxiv.org/abs/https://jamanetwork.com/journals/jamaneurology/articlepdf/592905/archneur\_51\_6\_015.pdf}{\tt
  arXiv:https://jamanetwork.com/journals/jamaneurology/articlepdf/592905/archneur\_51\_6\_015.pdf}.
%Type = Article
\bibitem[{Bertini et~al.(2022)Bertini, Allevi, Lutero, Calzà and
  Montesi}]{BERTINI2022101298}
\bibinfo{author}{Bertini, F.}, \bibinfo{author}{Allevi, D.},
  \bibinfo{author}{Lutero, G.}, \bibinfo{author}{Calzà, L.},
  \bibinfo{author}{Montesi, D.}, \bibinfo{year}{2022}.
\newblock \bibinfo{title}{An automatic alzheimer’s disease classifier based
  on spontaneous spoken english}.
\newblock \bibinfo{journal}{Computer Speech \& Language} \bibinfo{volume}{72},
  \bibinfo{pages}{101298}.
\newblock \URLprefix
  \url{https://www.sciencedirect.com/science/article/pii/S0885230821000991},
  \DOIprefix\doi{https://doi.org/10.1016/j.csl.2021.101298}.
%Type = Inproceedings
\bibitem[{Cai et~al.(2019)Cai, Cai and Wan}]{cai2019multi}
\bibinfo{author}{Cai, Y.}, \bibinfo{author}{Cai, H.}, \bibinfo{author}{Wan,
  X.}, \bibinfo{year}{2019}.
\newblock \bibinfo{title}{Multi-modal sarcasm detection in twitter with
  hierarchical fusion model}, in: \bibinfo{booktitle}{Proceedings of the 57th
  Annual Meeting of the Association for Computational Linguistics}, pp.
  \bibinfo{pages}{2506--2515}.
%Type = Article
\bibitem[{Chen et~al.(2022)Chen, Han and Chang}]{CHEN2022108980}
\bibinfo{author}{Chen, C.}, \bibinfo{author}{Han, D.}, \bibinfo{author}{Chang,
  C.C.}, \bibinfo{year}{2022}.
\newblock \bibinfo{title}{Caan: Context-aware attention network for visual
  question answering}.
\newblock \bibinfo{journal}{Pattern Recognition} \bibinfo{volume}{132},
  \bibinfo{pages}{108980}.
\newblock \URLprefix
  \url{https://www.sciencedirect.com/science/article/pii/S0031320322004605},
  \DOIprefix\doi{https://doi.org/10.1016/j.patcog.2022.108980}.
%Type = Inproceedings
\bibitem[{Chen et~al.(2021)Chen, Ye, Tang and Zhou}]{chen21r_interspeech}
\bibinfo{author}{Chen, J.}, \bibinfo{author}{Ye, J.}, \bibinfo{author}{Tang,
  F.}, \bibinfo{author}{Zhou, J.}, \bibinfo{year}{2021}.
\newblock \bibinfo{title}{{Automatic Detection of Alzheimer’s Disease Using
  Spontaneous Speech Only}}, in: \bibinfo{booktitle}{Proc. Interspeech 2021},
  pp. \bibinfo{pages}{3830--3834}.
\newblock \DOIprefix\doi{10.21437/Interspeech.2021-2002}.
%Type = Inproceedings
\bibitem[{Chollet(2017)}]{8099678}
\bibinfo{author}{Chollet, F.}, \bibinfo{year}{2017}.
\newblock \bibinfo{title}{Xception: Deep learning with depthwise separable
  convolutions}, in: \bibinfo{booktitle}{2017 IEEE Conference on Computer
  Vision and Pattern Recognition (CVPR)}, pp. \bibinfo{pages}{1800--1807}.
\newblock \DOIprefix\doi{10.1109/CVPR.2017.195}.
%Type = Article
\bibitem[{Courty et~al.(2017)Courty, Flamary, Tuia and Rakotomamonjy}]{7586038}
\bibinfo{author}{Courty, N.}, \bibinfo{author}{Flamary, R.},
  \bibinfo{author}{Tuia, D.}, \bibinfo{author}{Rakotomamonjy, A.},
  \bibinfo{year}{2017}.
\newblock \bibinfo{title}{Optimal transport for domain adaptation}.
\newblock \bibinfo{journal}{IEEE Transactions on Pattern Analysis and Machine
  Intelligence} \bibinfo{volume}{39}, \bibinfo{pages}{1853--1865}.
\newblock \DOIprefix\doi{10.1109/TPAMI.2016.2615921}.
%Type = Article
\bibitem[{Crowson et~al.(2016)Crowson, Atkinson and
  Therneau}]{doi:10.1177/0962280213497434}
\bibinfo{author}{Crowson, C.S.}, \bibinfo{author}{Atkinson, E.J.},
  \bibinfo{author}{Therneau, T.M.}, \bibinfo{year}{2016}.
\newblock \bibinfo{title}{Assessing calibration of prognostic risk scores}.
\newblock \bibinfo{journal}{Statistical Methods in Medical Research}
  \bibinfo{volume}{25}, \bibinfo{pages}{1692--1706}.
\newblock \URLprefix \url{https://doi.org/10.1177/0962280213497434},
  \DOIprefix\doi{10.1177/0962280213497434},
  \href{http://arxiv.org/abs/https://doi.org/10.1177/0962280213497434}{\tt
  arXiv:https://doi.org/10.1177/0962280213497434}. \bibinfo{note}{pMID:
  23907781}.
%Type = Inproceedings
\bibitem[{Cummins et~al.(2020)Cummins, Pan, Ren, Fritsch, Nallanthighal,
  Christensen, Blackburn, Schuller, Magimai-Doss, Strik
  et~al.}]{cummins2020comparison}
\bibinfo{author}{Cummins, N.}, \bibinfo{author}{Pan, Y.}, \bibinfo{author}{Ren,
  Z.}, \bibinfo{author}{Fritsch, J.}, \bibinfo{author}{Nallanthighal, V.S.},
  \bibinfo{author}{Christensen, H.}, \bibinfo{author}{Blackburn, D.},
  \bibinfo{author}{Schuller, B.W.}, \bibinfo{author}{Magimai-Doss, M.},
  \bibinfo{author}{Strik, H.}, et~al., \bibinfo{year}{2020}.
\newblock \bibinfo{title}{A comparison of acoustic and linguistics
  methodologies for alzheimer’s dementia recognition}, in:
  \bibinfo{booktitle}{Interspeech 2020},
  \bibinfo{organization}{ISCA-International Speech Communication Association}.
  pp. \bibinfo{pages}{2182--2186}.
%Type = Article
\bibitem[{Dawid(1982)}]{doi:10.1080/01621459.1982.10477856}
\bibinfo{author}{Dawid, A.P.}, \bibinfo{year}{1982}.
\newblock \bibinfo{title}{The well-calibrated bayesian}.
\newblock \bibinfo{journal}{Journal of the American Statistical Association}
  \bibinfo{volume}{77}, \bibinfo{pages}{605--610}.
\newblock \URLprefix
  \url{https://www.tandfonline.com/doi/abs/10.1080/01621459.1982.10477856},
  \DOIprefix\doi{10.1080/01621459.1982.10477856},
  \href{http://arxiv.org/abs/https://www.tandfonline.com/doi/pdf/10.1080/01621459.1982.10477856}{\tt
  arXiv:https://www.tandfonline.com/doi/pdf/10.1080/01621459.1982.10477856}.
%Type = Incollection
\bibitem[{Del~Barrio et~al.(2018)Del~Barrio, Cuesta-Albertos and
  Matr{\'a}n}]{del2018optimal}
\bibinfo{author}{Del~Barrio, E.}, \bibinfo{author}{Cuesta-Albertos, J.A.},
  \bibinfo{author}{Matr{\'a}n, C.}, \bibinfo{year}{2018}.
\newblock \bibinfo{title}{An optimal transportation approach for assessing
  almost stochastic order}, in: \bibinfo{booktitle}{The Mathematics of the
  Uncertain}. \bibinfo{publisher}{Springer}, pp. \bibinfo{pages}{33--44}.
%Type = Inproceedings
\bibitem[{Devlin et~al.(2019)Devlin, Chang, Lee and
  Toutanova}]{devlin-etal-2019-bert}
\bibinfo{author}{Devlin, J.}, \bibinfo{author}{Chang, M.W.},
  \bibinfo{author}{Lee, K.}, \bibinfo{author}{Toutanova, K.},
  \bibinfo{year}{2019}.
\newblock \bibinfo{title}{{BERT}: Pre-training of deep bidirectional
  transformers for language understanding}, in: \bibinfo{booktitle}{Proceedings
  of the 2019 Conference of the North {A}merican Chapter of the Association for
  Computational Linguistics: Human Language Technologies, Volume 1 (Long and
  Short Papers)}, \bibinfo{publisher}{Association for Computational
  Linguistics}, \bibinfo{address}{Minneapolis, Minnesota}. pp.
  \bibinfo{pages}{4171--4186}.
\newblock \URLprefix \url{https://aclanthology.org/N19-1423},
  \DOIprefix\doi{10.18653/v1/N19-1423}.
%Type = Inproceedings
\bibitem[{Dror et~al.(2019)Dror, Shlomov and Reichart}]{dror2019deep}
\bibinfo{author}{Dror, R.}, \bibinfo{author}{Shlomov, S.},
  \bibinfo{author}{Reichart, R.}, \bibinfo{year}{2019}.
\newblock \bibinfo{title}{Deep dominance - how to properly compare deep neural
  models}, in: \bibinfo{editor}{Korhonen, A.}, \bibinfo{editor}{Traum, D.R.},
  \bibinfo{editor}{M{\`{a}}rquez, L.} (Eds.), \bibinfo{booktitle}{Proceedings
  of the 57th Conference of the Association for Computational Linguistics,
  {ACL} 2019, Florence, Italy, July 28- August 2, 2019, Volume 1: Long Papers},
  \bibinfo{publisher}{Association for Computational Linguistics}. pp.
  \bibinfo{pages}{2773--2785}.
\newblock \URLprefix \url{https://doi.org/10.18653/v1/p19-1266},
  \DOIprefix\doi{10.18653/v1/p19-1266}.
%Type = Inproceedings
\bibitem[{Edwards et~al.(2020)Edwards, Dognin, Bollepalli and
  Singh}]{edwards20_interspeech}
\bibinfo{author}{Edwards, E.}, \bibinfo{author}{Dognin, C.},
  \bibinfo{author}{Bollepalli, B.}, \bibinfo{author}{Singh, M.},
  \bibinfo{year}{2020}.
\newblock \bibinfo{title}{{Multiscale System for Alzheimer’s Dementia
  Recognition Through Spontaneous Speech}}, in: \bibinfo{booktitle}{Proc.
  Interspeech 2020}, pp. \bibinfo{pages}{2197--2201}.
\newblock \DOIprefix\doi{10.21437/Interspeech.2020-2781}.
%Type = Article
\bibitem[{Ferradans et~al.(2014)Ferradans, Papadakis, Peyr{\'e} and
  Aujol}]{ferradans2014regularized}
\bibinfo{author}{Ferradans, S.}, \bibinfo{author}{Papadakis, N.},
  \bibinfo{author}{Peyr{\'e}, G.}, \bibinfo{author}{Aujol, J.F.},
  \bibinfo{year}{2014}.
\newblock \bibinfo{title}{Regularized discrete optimal transport}.
\newblock \bibinfo{journal}{SIAM Journal on Imaging Sciences}
  \bibinfo{volume}{7}, \bibinfo{pages}{1853--1882}.
%Type = Article
\bibitem[{Flamary et~al.(2021)Flamary, Courty, Gramfort, Alaya, Boisbunon,
  Chambon, Chapel, Corenflos, Fatras, Fournier, Gautheron, Gayraud, Janati,
  Rakotomamonjy, Redko, Rolet, Schutz, Seguy, Sutherland, Tavenard, Tong and
  Vayer}]{flamary2021pot}
\bibinfo{author}{Flamary, R.}, \bibinfo{author}{Courty, N.},
  \bibinfo{author}{Gramfort, A.}, \bibinfo{author}{Alaya, M.Z.},
  \bibinfo{author}{Boisbunon, A.}, \bibinfo{author}{Chambon, S.},
  \bibinfo{author}{Chapel, L.}, \bibinfo{author}{Corenflos, A.},
  \bibinfo{author}{Fatras, K.}, \bibinfo{author}{Fournier, N.},
  \bibinfo{author}{Gautheron, L.}, \bibinfo{author}{Gayraud, N.T.},
  \bibinfo{author}{Janati, H.}, \bibinfo{author}{Rakotomamonjy, A.},
  \bibinfo{author}{Redko, I.}, \bibinfo{author}{Rolet, A.},
  \bibinfo{author}{Schutz, A.}, \bibinfo{author}{Seguy, V.},
  \bibinfo{author}{Sutherland, D.J.}, \bibinfo{author}{Tavenard, R.},
  \bibinfo{author}{Tong, A.}, \bibinfo{author}{Vayer, T.},
  \bibinfo{year}{2021}.
\newblock \bibinfo{title}{Pot: Python optimal transport}.
\newblock \bibinfo{journal}{Journal of Machine Learning Research}
  \bibinfo{volume}{22}, \bibinfo{pages}{1--8}.
\newblock \URLprefix \url{http://jmlr.org/papers/v22/20-451.html}.
%Type = Article
\bibitem[{Freitag et~al.(2017)Freitag, Amiriparian, Pugachevskiy, Cummins and
  Schuller}]{freitag2017audeep}
\bibinfo{author}{Freitag, M.}, \bibinfo{author}{Amiriparian, S.},
  \bibinfo{author}{Pugachevskiy, S.}, \bibinfo{author}{Cummins, N.},
  \bibinfo{author}{Schuller, B.}, \bibinfo{year}{2017}.
\newblock \bibinfo{title}{audeep: Unsupervised learning of representations from
  audio with deep recurrent neural networks}.
\newblock \bibinfo{journal}{The Journal of Machine Learning Research}
  \bibinfo{volume}{18}, \bibinfo{pages}{6340--6344}.
%Type = Inproceedings
\bibitem[{Gu et~al.(2018)Gu, Yang, Fu, Chen, Li and
  Marsic}]{gu-etal-2018-hybrid}
\bibinfo{author}{Gu, Y.}, \bibinfo{author}{Yang, K.}, \bibinfo{author}{Fu, S.},
  \bibinfo{author}{Chen, S.}, \bibinfo{author}{Li, X.},
  \bibinfo{author}{Marsic, I.}, \bibinfo{year}{2018}.
\newblock \bibinfo{title}{Hybrid attention based multimodal network for spoken
  language classification}, in: \bibinfo{booktitle}{Proceedings of the 27th
  International Conference on Computational Linguistics},
  \bibinfo{publisher}{Association for Computational Linguistics},
  \bibinfo{address}{Santa Fe, New Mexico, USA}. pp.
  \bibinfo{pages}{2379--2390}.
\newblock \URLprefix \url{https://aclanthology.org/C18-1201}.
%Type = Inproceedings
\bibitem[{Guo et~al.(2017)Guo, Pleiss, Sun and Weinberger}]{pmlr-v70-guo17a}
\bibinfo{author}{Guo, C.}, \bibinfo{author}{Pleiss, G.}, \bibinfo{author}{Sun,
  Y.}, \bibinfo{author}{Weinberger, K.Q.}, \bibinfo{year}{2017}.
\newblock \bibinfo{title}{On calibration of modern neural networks}, in:
  \bibinfo{editor}{Precup, D.}, \bibinfo{editor}{Teh, Y.W.} (Eds.),
  \bibinfo{booktitle}{Proceedings of the 34th International Conference on
  Machine Learning}, \bibinfo{publisher}{PMLR}. pp.
  \bibinfo{pages}{1321--1330}.
\newblock \URLprefix \url{https://proceedings.mlr.press/v70/guo17a.html}.
%Type = Article
\bibitem[{Haulcy and Glass(2021)}]{10.3389/fpsyg.2020.624137}
\bibinfo{author}{Haulcy, R.}, \bibinfo{author}{Glass, J.},
  \bibinfo{year}{2021}.
\newblock \bibinfo{title}{Classifying alzheimer's disease using audio and
  text-based representations of speech}.
\newblock \bibinfo{journal}{Frontiers in Psychology} \bibinfo{volume}{11},
  \bibinfo{pages}{3833}.
\newblock \URLprefix
  \url{https://www.frontiersin.org/article/10.3389/fpsyg.2020.624137},
  \DOIprefix\doi{10.3389/fpsyg.2020.624137}.
%Type = Article
\bibitem[{Ilias and Askounis(2022a)}]{9769980}
\bibinfo{author}{Ilias, L.}, \bibinfo{author}{Askounis, D.},
  \bibinfo{year}{2022}a.
\newblock \bibinfo{title}{Explainable identification of dementia from
  transcripts using transformer networks}.
\newblock \bibinfo{journal}{IEEE Journal of Biomedical and Health Informatics}
  \bibinfo{volume}{26}, \bibinfo{pages}{4153--4164}.
\newblock \DOIprefix\doi{10.1109/JBHI.2022.3172479}.
%Type = Article
\bibitem[{Ilias and Askounis(2022b)}]{10.3389/fnagi.2022.830943}
\bibinfo{author}{Ilias, L.}, \bibinfo{author}{Askounis, D.},
  \bibinfo{year}{2022}b.
\newblock \bibinfo{title}{Multimodal deep learning models for detecting
  dementia from speech and transcripts}.
\newblock \bibinfo{journal}{Frontiers in Aging Neuroscience}
  \bibinfo{volume}{14}.
\newblock \URLprefix
  \url{https://www.frontiersin.org/articles/10.3389/fnagi.2022.830943},
  \DOIprefix\doi{10.3389/fnagi.2022.830943}.
%Type = Inproceedings
\bibitem[{Ilias et~al.(2022)Ilias, Askounis and Psarras}]{9926818}
\bibinfo{author}{Ilias, L.}, \bibinfo{author}{Askounis, D.},
  \bibinfo{author}{Psarras, J.}, \bibinfo{year}{2022}.
\newblock \bibinfo{title}{A multimodal approach for dementia detection from
  spontaneous speech with tensor fusion layer}, in: \bibinfo{booktitle}{2022
  IEEE-EMBS International Conference on Biomedical and Health Informatics
  (BHI)}, pp. \bibinfo{pages}{1--5}.
\newblock \DOIprefix\doi{10.1109/BHI56158.2022.9926818}.
%Type = Article
\bibitem[{Ilias et~al.(2023)Ilias, Askounis and Psarras}]{ILIAS2023101485}
\bibinfo{author}{Ilias, L.}, \bibinfo{author}{Askounis, D.},
  \bibinfo{author}{Psarras, J.}, \bibinfo{year}{2023}.
\newblock \bibinfo{title}{Detecting dementia from speech and transcripts using
  transformers}.
\newblock \bibinfo{journal}{Computer Speech \& Language} \bibinfo{volume}{79},
  \bibinfo{pages}{101485}.
\newblock \URLprefix
  \url{https://www.sciencedirect.com/science/article/pii/S0885230823000049},
  \DOIprefix\doi{https://doi.org/10.1016/j.csl.2023.101485}.
%Type = Article
\bibitem[{Jiang et~al.(2011)Jiang, Osl, Kim and
  Ohno-Machado}]{10.1136/amiajnl-2011-000291}
\bibinfo{author}{Jiang, X.}, \bibinfo{author}{Osl, M.}, \bibinfo{author}{Kim,
  J.}, \bibinfo{author}{Ohno-Machado, L.}, \bibinfo{year}{2011}.
\newblock \bibinfo{title}{{Calibrating predictive model estimates to support
  personalized medicine}}.
\newblock \bibinfo{journal}{Journal of the American Medical Informatics
  Association} \bibinfo{volume}{19}, \bibinfo{pages}{263--274}.
\newblock \URLprefix \url{https://doi.org/10.1136/amiajnl-2011-000291},
  \DOIprefix\doi{10.1136/amiajnl-2011-000291},
  \href{http://arxiv.org/abs/https://academic.oup.com/jamia/article-pdf/19/2/263/17374049/19-2-263.pdf}{\tt
  arXiv:https://academic.oup.com/jamia/article-pdf/19/2/263/17374049/19-2-263.pdf}.
%Type = Inproceedings
\bibitem[{Karlekar et~al.(2018)Karlekar, Niu and
  Bansal}]{karlekar-etal-2018-detecting}
\bibinfo{author}{Karlekar, S.}, \bibinfo{author}{Niu, T.},
  \bibinfo{author}{Bansal, M.}, \bibinfo{year}{2018}.
\newblock \bibinfo{title}{Detecting linguistic characteristics of
  {A}lzheimer{'}s dementia by interpreting neural models}, in:
  \bibinfo{booktitle}{Proceedings of the 2018 Conference of the North
  {A}merican Chapter of the Association for Computational Linguistics: Human
  Language Technologies, Volume 2 (Short Papers)},
  \bibinfo{publisher}{Association for Computational Linguistics},
  \bibinfo{address}{New Orleans, Louisiana}. pp. \bibinfo{pages}{701--707}.
\newblock \URLprefix \url{https://aclanthology.org/N18-2110},
  \DOIprefix\doi{10.18653/v1/N18-2110}.
%Type = Inproceedings
\bibitem[{Khodabakhsh et~al.(2014)Khodabakhsh, Kuşxuoğlu and
  Demiroğlu}]{6864431}
\bibinfo{author}{Khodabakhsh, A.}, \bibinfo{author}{Kuşxuoğlu, S.},
  \bibinfo{author}{Demiroğlu, C.}, \bibinfo{year}{2014}.
\newblock \bibinfo{title}{Natural language features for detection of
  alzheimer's disease in conversational speech}, in:
  \bibinfo{booktitle}{IEEE-EMBS International Conference on Biomedical and
  Health Informatics (BHI)}, pp. \bibinfo{pages}{581--584}.
\newblock \DOIprefix\doi{10.1109/BHI.2014.6864431}.
%Type = Inproceedings
\bibitem[{Koo et~al.(2020)Koo, Lee, Pyo, Jo and Lee}]{koo20_interspeech}
\bibinfo{author}{Koo, J.}, \bibinfo{author}{Lee, J.H.}, \bibinfo{author}{Pyo,
  J.}, \bibinfo{author}{Jo, Y.}, \bibinfo{author}{Lee, K.},
  \bibinfo{year}{2020}.
\newblock \bibinfo{title}{{Exploiting Multi-Modal Features from Pre-Trained
  Networks for Alzheimer’s Dementia Recognition}}, in:
  \bibinfo{booktitle}{Proc. Interspeech 2020}, pp. \bibinfo{pages}{2217--2221}.
\newblock \DOIprefix\doi{10.21437/Interspeech.2020-3153}.
%Type = Inproceedings
\bibitem[{Kumar et~al.(2011)Kumar, Kim and Stern}]{5947425}
\bibinfo{author}{Kumar, K.}, \bibinfo{author}{Kim, C.}, \bibinfo{author}{Stern,
  R.M.}, \bibinfo{year}{2011}.
\newblock \bibinfo{title}{Delta-spectral cepstral coefficients for robust
  speech recognition}, in: \bibinfo{booktitle}{2011 IEEE International
  Conference on Acoustics, Speech and Signal Processing (ICASSP)}, pp.
  \bibinfo{pages}{4784--4787}.
\newblock \DOIprefix\doi{10.1109/ICASSP.2011.5947425}.
%Type = Techreport
\bibitem[{Lee et~al.(2016)Lee, Burkholder, Flinn and
  Coppess}]{lee-et-al-pylangacq:2016}
\bibinfo{author}{Lee, J.L.}, \bibinfo{author}{Burkholder, R.},
  \bibinfo{author}{Flinn, G.B.}, \bibinfo{author}{Coppess, E.R.},
  \bibinfo{year}{2016}.
\newblock \bibinfo{title}{Working with CHAT transcripts in Python}.
\newblock \bibinfo{type}{Technical Report} \bibinfo{number}{TR-2016-02}.
  Department of Computer Science, University of Chicago.
%Type = Inproceedings
\bibitem[{Lu et~al.(2016)Lu, Yang, Batra and Parikh}]{10.5555/3157096.3157129}
\bibinfo{author}{Lu, J.}, \bibinfo{author}{Yang, J.}, \bibinfo{author}{Batra,
  D.}, \bibinfo{author}{Parikh, D.}, \bibinfo{year}{2016}.
\newblock \bibinfo{title}{Hierarchical question-image co-attention for visual
  question answering}, in: \bibinfo{booktitle}{Proceedings of the 30th
  International Conference on Neural Information Processing Systems},
  \bibinfo{publisher}{Curran Associates Inc.}, \bibinfo{address}{Red Hook, NY,
  USA}. p. \bibinfo{pages}{289–297}.
%Type = Article
\bibitem[{Lu et~al.(2021)Lu, Wang and Zhang}]{lu2021detection}
\bibinfo{author}{Lu, S.}, \bibinfo{author}{Wang, S.H.}, \bibinfo{author}{Zhang,
  Y.D.}, \bibinfo{year}{2021}.
\newblock \bibinfo{title}{Detection of abnormal brain in mri via improved
  alexnet and elm optimized by chaotic bat algorithm}.
\newblock \bibinfo{journal}{Neural Computing and Applications}
  \bibinfo{volume}{33}, \bibinfo{pages}{10799--10811}.
%Type = Article
\bibitem[{Lu et~al.(2022)Lu, Zhu, Gorriz, Wang and
  Zhang}]{https://doi.org/10.1002/int.22686}
\bibinfo{author}{Lu, S.}, \bibinfo{author}{Zhu, Z.}, \bibinfo{author}{Gorriz,
  J.M.}, \bibinfo{author}{Wang, S.H.}, \bibinfo{author}{Zhang, Y.D.},
  \bibinfo{year}{2022}.
\newblock \bibinfo{title}{Nagnn: Classification of covid-19 based on
  neighboring aware representation from deep graph neural network}.
\newblock \bibinfo{journal}{International Journal of Intelligent Systems}
  \bibinfo{volume}{37}, \bibinfo{pages}{1572--1598}.
\newblock \URLprefix
  \url{https://onlinelibrary.wiley.com/doi/abs/10.1002/int.22686},
  \DOIprefix\doi{https://doi.org/10.1002/int.22686},
  \href{http://arxiv.org/abs/https://onlinelibrary.wiley.com/doi/pdf/10.1002/int.22686}{\tt
  arXiv:https://onlinelibrary.wiley.com/doi/pdf/10.1002/int.22686}.
%Type = Article
\bibitem[{Luz et~al.(2023)Luz, Haider, Fromm, Lazarou, Kompatsiaris and
  MacWhinney}]{luz2023multilingual}
\bibinfo{author}{Luz, S.}, \bibinfo{author}{Haider, F.},
  \bibinfo{author}{Fromm, D.}, \bibinfo{author}{Lazarou, I.},
  \bibinfo{author}{Kompatsiaris, I.}, \bibinfo{author}{MacWhinney, B.},
  \bibinfo{year}{2023}.
\newblock \bibinfo{title}{Multilingual alzheimer's dementia recognition through
  spontaneous speech: a signal processing grand challenge}.
\newblock \bibinfo{journal}{arXiv preprint arXiv:2301.05562} .
%Type = Inproceedings
\bibitem[{Luz et~al.(2020)Luz, Haider, de~la Fuente, Fromm and
  MacWhinney}]{luz20_interspeech}
\bibinfo{author}{Luz, S.}, \bibinfo{author}{Haider, F.}, \bibinfo{author}{de~la
  Fuente, S.}, \bibinfo{author}{Fromm, D.}, \bibinfo{author}{MacWhinney, B.},
  \bibinfo{year}{2020}.
\newblock \bibinfo{title}{{Alzheimer’s Dementia Recognition Through
  Spontaneous Speech: The ADReSS Challenge}}, in: \bibinfo{booktitle}{Proc.
  Interspeech 2020}, pp. \bibinfo{pages}{2172--2176}.
\newblock \DOIprefix\doi{10.21437/Interspeech.2020-2571}.
%Type = Inproceedings
\bibitem[{Luz et~al.(2021)Luz, Haider, de~la Fuente, Fromm and
  MacWhinney}]{luz21_interspeech}
\bibinfo{author}{Luz, S.}, \bibinfo{author}{Haider, F.}, \bibinfo{author}{de~la
  Fuente, S.}, \bibinfo{author}{Fromm, D.}, \bibinfo{author}{MacWhinney, B.},
  \bibinfo{year}{2021}.
\newblock \bibinfo{title}{{Detecting Cognitive Decline Using Speech Only: The
  ADReSSo Challenge}}, in: \bibinfo{booktitle}{Proc. Interspeech 2021}, pp.
  \bibinfo{pages}{3780--3784}.
\newblock \DOIprefix\doi{10.21437/Interspeech.2021-1220}.
%Type = Article
\bibitem[{MacWhinney(2000)}]{10.1162/coli.2000.26.4.657}
\bibinfo{author}{MacWhinney, B.}, \bibinfo{year}{2000}.
\newblock \bibinfo{title}{{The CHILDES Project: Tools for Analyzing Talk (third
  edition): Volume I: Transcription format and programs, Volume II: The
  database}}.
\newblock \bibinfo{journal}{Computational Linguistics} \bibinfo{volume}{26},
  \bibinfo{pages}{657--657}.
\newblock \URLprefix \url{https://doi.org/10.1162/coli.2000.26.4.657},
  \DOIprefix\doi{10.1162/coli.2000.26.4.657},
  \href{http://arxiv.org/abs/https://direct.mit.edu/coli/article-pdf/26/4/657/1797615/coli.2000.26.4.657.pdf}{\tt
  arXiv:https://direct.mit.edu/coli/article-pdf/26/4/657/1797615/coli.2000.26.4.657.pdf}.
%Type = Article
\bibitem[{Mahajan and Baths(2021)}]{10.3389/fnagi.2021.623607}
\bibinfo{author}{Mahajan, P.}, \bibinfo{author}{Baths, V.},
  \bibinfo{year}{2021}.
\newblock \bibinfo{title}{Acoustic and language based deep learning approaches
  for alzheimer's dementia detection from spontaneous speech}.
\newblock \bibinfo{journal}{Frontiers in Aging Neuroscience}
  \bibinfo{volume}{13}, \bibinfo{pages}{20}.
\newblock \URLprefix
  \url{https://www.frontiersin.org/article/10.3389/fnagi.2021.623607},
  \DOIprefix\doi{10.3389/fnagi.2021.623607}.
%Type = Inproceedings
\bibitem[{Martinc and Pollak(2020)}]{martinc20_interspeech}
\bibinfo{author}{Martinc, M.}, \bibinfo{author}{Pollak, S.},
  \bibinfo{year}{2020}.
\newblock \bibinfo{title}{{Tackling the ADReSS Challenge: A Multimodal Approach
  to the Automated Recognition of Alzheimer’s Dementia}}, in:
  \bibinfo{booktitle}{Proc. Interspeech 2020}, pp. \bibinfo{pages}{2157--2161}.
\newblock \DOIprefix\doi{10.21437/Interspeech.2020-2202}.
%Type = Misc
\bibitem[{McFee et~al.(2022)McFee, Metsai, McVicar, Balke, Thomé, Raffel,
  Zalkow, Malek, Dana, Lee, Nieto, Ellis, Mason, Battenberg, Seyfarth,
  Yamamoto, viktorandreevichmorozov, Choi, Moore, Bittner, Hidaka, Wei,
  nullmightybofo, Weiss, Hereñú, Stöter, Nickel, Friesch, Vollrath and
  Kim}]{brian_mcfee_2022_6759664}
\bibinfo{author}{McFee, B.}, \bibinfo{author}{Metsai, A.},
  \bibinfo{author}{McVicar, M.}, \bibinfo{author}{Balke, S.},
  \bibinfo{author}{Thomé, C.}, \bibinfo{author}{Raffel, C.},
  \bibinfo{author}{Zalkow, F.}, \bibinfo{author}{Malek, A.},
  \bibinfo{author}{Dana}, \bibinfo{author}{Lee, K.}, \bibinfo{author}{Nieto,
  O.}, \bibinfo{author}{Ellis, D.}, \bibinfo{author}{Mason, J.},
  \bibinfo{author}{Battenberg, E.}, \bibinfo{author}{Seyfarth, S.},
  \bibinfo{author}{Yamamoto, R.}, \bibinfo{author}{viktorandreevichmorozov},
  \bibinfo{author}{Choi, K.}, \bibinfo{author}{Moore, J.},
  \bibinfo{author}{Bittner, R.}, \bibinfo{author}{Hidaka, S.},
  \bibinfo{author}{Wei, Z.}, \bibinfo{author}{nullmightybofo},
  \bibinfo{author}{Weiss, A.}, \bibinfo{author}{Hereñú, D.},
  \bibinfo{author}{Stöter, F.R.}, \bibinfo{author}{Nickel, L.},
  \bibinfo{author}{Friesch, P.}, \bibinfo{author}{Vollrath, M.},
  \bibinfo{author}{Kim, T.}, \bibinfo{year}{2022}.
\newblock \bibinfo{title}{librosa/librosa: 0.9.2}.
\newblock \URLprefix \url{https://doi.org/10.5281/zenodo.6759664},
  \DOIprefix\doi{10.5281/zenodo.6759664}.
%Type = Inproceedings
\bibitem[{McFee et~al.(2015)McFee, Raffel, Liang, Ellis, McVicar, Battenberg
  and Nieto}]{mcfee2015librosa}
\bibinfo{author}{McFee, B.}, \bibinfo{author}{Raffel, C.},
  \bibinfo{author}{Liang, D.}, \bibinfo{author}{Ellis, D.P.},
  \bibinfo{author}{McVicar, M.}, \bibinfo{author}{Battenberg, E.},
  \bibinfo{author}{Nieto, O.}, \bibinfo{year}{2015}.
\newblock \bibinfo{title}{librosa: Audio and music signal analysis in python},
  in: \bibinfo{booktitle}{Proceedings of the 14th python in science
  conference}, pp. \bibinfo{pages}{18--25}.
%Type = Inproceedings
\bibitem[{Mialon et~al.(2021)Mialon, Chen, d'Aspremont and
  Mairal}]{mialon2021a}
\bibinfo{author}{Mialon, G.}, \bibinfo{author}{Chen, D.},
  \bibinfo{author}{d'Aspremont, A.}, \bibinfo{author}{Mairal, J.},
  \bibinfo{year}{2021}.
\newblock \bibinfo{title}{A trainable optimal transport embedding for feature
  aggregation and its relationship to attention}, in:
  \bibinfo{booktitle}{International Conference on Learning Representations}.
%Type = Misc
\bibitem[{Mittal et~al.(2021)Mittal, Sahoo, Datar, Kadiwala, Shalu and
  Mathew}]{mittal2021multimodal}
\bibinfo{author}{Mittal, A.}, \bibinfo{author}{Sahoo, S.},
  \bibinfo{author}{Datar, A.}, \bibinfo{author}{Kadiwala, J.},
  \bibinfo{author}{Shalu, H.}, \bibinfo{author}{Mathew, J.},
  \bibinfo{year}{2021}.
\newblock \bibinfo{title}{Multi-modal detection of alzheimer's disease from
  speech and text}.
\newblock \href{http://arxiv.org/abs/2012.00096}{\tt arXiv:2012.00096}.
%Type = Inproceedings
\bibitem[{M\"{u}ller et~al.(2019)M\"{u}ller, Kornblith and
  Hinton}]{NEURIPS2019_f1748d6b}
\bibinfo{author}{M\"{u}ller, R.}, \bibinfo{author}{Kornblith, S.},
  \bibinfo{author}{Hinton, G.E.}, \bibinfo{year}{2019}.
\newblock \bibinfo{title}{When does label smoothing help?}, in:
  \bibinfo{editor}{Wallach, H.}, \bibinfo{editor}{Larochelle, H.},
  \bibinfo{editor}{Beygelzimer, A.}, \bibinfo{editor}{d\textquotesingle
  Alch\'{e}-Buc, F.}, \bibinfo{editor}{Fox, E.}, \bibinfo{editor}{Garnett, R.}
  (Eds.), \bibinfo{booktitle}{Advances in Neural Information Processing
  Systems}, \bibinfo{publisher}{Curran Associates, Inc.}
\newblock \URLprefix
  \url{https://proceedings.neurips.cc/paper/2019/file/f1748d6b0fd9d439f71450117eba2725-Paper.pdf}.
%Type = Article
\bibitem[{Murphy and
  Epstein(1967)}]{VerificationofProbabilisticPredictionsABriefReview}
\bibinfo{author}{Murphy, A.H.}, \bibinfo{author}{Epstein, E.S.},
  \bibinfo{year}{1967}.
\newblock \bibinfo{title}{Verification of probabilistic predictions: A brief
  review}.
\newblock \bibinfo{journal}{Journal of Applied Meteorology and Climatology}
  \bibinfo{volume}{6}, \bibinfo{pages}{748 -- 755}.
\newblock \URLprefix
  \url{https://journals.ametsoc.org/view/journals/apme/6/5/1520-0450_1967_006_0748_voppab_2_0_co_2.xml},
  \DOIprefix\doi{10.1175/1520-0450(1967)006<0748:VOPPAB>2.0.CO;2}.
%Type = Inproceedings
\bibitem[{Naeini et~al.(2015)Naeini, Cooper and
  Hauskrecht}]{naeini2015obtaining}
\bibinfo{author}{Naeini, M.P.}, \bibinfo{author}{Cooper, G.},
  \bibinfo{author}{Hauskrecht, M.}, \bibinfo{year}{2015}.
\newblock \bibinfo{title}{Obtaining well calibrated probabilities using
  bayesian binning}, in: \bibinfo{booktitle}{Twenty-Ninth AAAI Conference on
  Artificial Intelligence}.
%Type = Inproceedings
\bibitem[{Nixon et~al.(2019)Nixon, Dusenberry, Zhang, Jerfel and
  Tran}]{Nixon_2019_CVPR_Workshops}
\bibinfo{author}{Nixon, J.}, \bibinfo{author}{Dusenberry, M.W.},
  \bibinfo{author}{Zhang, L.}, \bibinfo{author}{Jerfel, G.},
  \bibinfo{author}{Tran, D.}, \bibinfo{year}{2019}.
\newblock \bibinfo{title}{Measuring calibration in deep learning}, in:
  \bibinfo{booktitle}{Proceedings of the IEEE/CVF Conference on Computer Vision
  and Pattern Recognition (CVPR) Workshops}.
%Type = Inproceedings
\bibitem[{Pan et~al.(2020)Pan, Lin, Fu, Qi and Wang}]{pan-etal-2020-modeling}
\bibinfo{author}{Pan, H.}, \bibinfo{author}{Lin, Z.}, \bibinfo{author}{Fu, P.},
  \bibinfo{author}{Qi, Y.}, \bibinfo{author}{Wang, W.}, \bibinfo{year}{2020}.
\newblock \bibinfo{title}{Modeling intra and inter-modality incongruity for
  multi-modal sarcasm detection}, in: \bibinfo{booktitle}{Findings of the
  Association for Computational Linguistics: EMNLP 2020},
  \bibinfo{publisher}{Association for Computational Linguistics},
  \bibinfo{address}{Online}. pp. \bibinfo{pages}{1383--1392}.
\newblock \URLprefix \url{https://aclanthology.org/2020.findings-emnlp.124},
  \DOIprefix\doi{10.18653/v1/2020.findings-emnlp.124}.
%Type = Inproceedings
\bibitem[{Pan et~al.(2021)Pan, Mirheidari, Harris, Thompson, Jones, Snowden,
  Blackburn and Christensen}]{pan21c_interspeech}
\bibinfo{author}{Pan, Y.}, \bibinfo{author}{Mirheidari, B.},
  \bibinfo{author}{Harris, J.M.}, \bibinfo{author}{Thompson, J.C.},
  \bibinfo{author}{Jones, M.}, \bibinfo{author}{Snowden, J.S.},
  \bibinfo{author}{Blackburn, D.}, \bibinfo{author}{Christensen, H.},
  \bibinfo{year}{2021}.
\newblock \bibinfo{title}{{Using the Outputs of Different Automatic Speech
  Recognition Paradigms for Acoustic- and BERT-Based Alzheimer’s Dementia
  Detection Through Spontaneous Speech}}, in: \bibinfo{booktitle}{Proc.
  Interspeech 2021}, pp. \bibinfo{pages}{3810--3814}.
\newblock \DOIprefix\doi{10.21437/Interspeech.2021-1519}.
%Type = Inproceedings
\bibitem[{Pappagari et~al.(2021)Pappagari, Cho, Joshi, Moro-Velázquez,
  Żelasko, Villalba and Dehak}]{pappagari21_interspeech}
\bibinfo{author}{Pappagari, R.}, \bibinfo{author}{Cho, J.},
  \bibinfo{author}{Joshi, S.}, \bibinfo{author}{Moro-Velázquez, L.},
  \bibinfo{author}{Żelasko, P.}, \bibinfo{author}{Villalba, J.},
  \bibinfo{author}{Dehak, N.}, \bibinfo{year}{2021}.
\newblock \bibinfo{title}{{Automatic Detection and Assessment of Alzheimer
  Disease Using Speech and Language Technologies in Low-Resource Scenarios}},
  in: \bibinfo{booktitle}{Proc. Interspeech 2021}, pp.
  \bibinfo{pages}{3825--3829}.
\newblock \DOIprefix\doi{10.21437/Interspeech.2021-1850}.
%Type = Inproceedings
\bibitem[{Pappagari et~al.(2020)Pappagari, Cho, Moro-Velázquez and
  Dehak}]{pappagari20_interspeech}
\bibinfo{author}{Pappagari, R.}, \bibinfo{author}{Cho, J.},
  \bibinfo{author}{Moro-Velázquez, L.}, \bibinfo{author}{Dehak, N.},
  \bibinfo{year}{2020}.
\newblock \bibinfo{title}{{Using State of the Art Speaker Recognition and
  Natural Language Processing Technologies to Detect Alzheimer’s Disease and
  Assess its Severity}}, in: \bibinfo{booktitle}{Proc. Interspeech 2020}, pp.
  \bibinfo{pages}{2177--2181}.
\newblock \DOIprefix\doi{10.21437/Interspeech.2020-2587}.
%Type = Inproceedings
\bibitem[{Park et~al.(2019)Park, Chan, Zhang, Chiu, Zoph, Cubuk and
  Le}]{park19e_interspeech}
\bibinfo{author}{Park, D.S.}, \bibinfo{author}{Chan, W.},
  \bibinfo{author}{Zhang, Y.}, \bibinfo{author}{Chiu, C.C.},
  \bibinfo{author}{Zoph, B.}, \bibinfo{author}{Cubuk, E.D.},
  \bibinfo{author}{Le, Q.V.}, \bibinfo{year}{2019}.
\newblock \bibinfo{title}{{SpecAugment: A Simple Data Augmentation Method for
  Automatic Speech Recognition}}, in: \bibinfo{booktitle}{Proc. Interspeech
  2019}, pp. \bibinfo{pages}{2613--2617}.
\newblock \DOIprefix\doi{10.21437/Interspeech.2019-2680}.
%Type = Incollection
\bibitem[{Paszke et~al.(2019)Paszke, Gross, Massa, Lerer, Bradbury, Chanan,
  Killeen, Lin, Gimelshein, Antiga, Desmaison, Kopf, Yang, DeVito, Raison,
  Tejani, Chilamkurthy, Steiner, Fang, Bai and Chintala}]{NEURIPS2019_9015}
\bibinfo{author}{Paszke, A.}, \bibinfo{author}{Gross, S.},
  \bibinfo{author}{Massa, F.}, \bibinfo{author}{Lerer, A.},
  \bibinfo{author}{Bradbury, J.}, \bibinfo{author}{Chanan, G.},
  \bibinfo{author}{Killeen, T.}, \bibinfo{author}{Lin, Z.},
  \bibinfo{author}{Gimelshein, N.}, \bibinfo{author}{Antiga, L.},
  \bibinfo{author}{Desmaison, A.}, \bibinfo{author}{Kopf, A.},
  \bibinfo{author}{Yang, E.}, \bibinfo{author}{DeVito, Z.},
  \bibinfo{author}{Raison, M.}, \bibinfo{author}{Tejani, A.},
  \bibinfo{author}{Chilamkurthy, S.}, \bibinfo{author}{Steiner, B.},
  \bibinfo{author}{Fang, L.}, \bibinfo{author}{Bai, J.},
  \bibinfo{author}{Chintala, S.}, \bibinfo{year}{2019}.
\newblock \bibinfo{title}{Pytorch: An imperative style, high-performance deep
  learning library}, in: \bibinfo{editor}{Wallach, H.},
  \bibinfo{editor}{Larochelle, H.}, \bibinfo{editor}{Beygelzimer, A.},
  \bibinfo{editor}{d\textquotesingle Alch\'{e}-Buc, F.}, \bibinfo{editor}{Fox,
  E.}, \bibinfo{editor}{Garnett, R.} (Eds.), \bibinfo{booktitle}{Advances in
  Neural Information Processing Systems 32}. \bibinfo{publisher}{Curran
  Associates, Inc.}, pp. \bibinfo{pages}{8024--8035}.
\newblock \URLprefix
  \url{http://papers.neurips.cc/paper/9015-pytorch-an-imperative-style-high-performance-deep-learning-library.pdf}.
%Type = Inproceedings
\bibitem[{Pompili et~al.(2020)Pompili, Rolland and
  Abad}]{pompili20_interspeech}
\bibinfo{author}{Pompili, A.}, \bibinfo{author}{Rolland, T.},
  \bibinfo{author}{Abad, A.}, \bibinfo{year}{2020}.
\newblock \bibinfo{title}{{The INESC-ID Multi-Modal System for the ADReSS 2020
  Challenge}}, in: \bibinfo{booktitle}{Proc. Interspeech 2020}, pp.
  \bibinfo{pages}{2202--2206}.
\newblock \DOIprefix\doi{10.21437/Interspeech.2020-2833}.
%Type = Inproceedings
\bibitem[{Pramanick et~al.(2022)Pramanick, Roy and Patel}]{Pramanick_2022_WACV}
\bibinfo{author}{Pramanick, S.}, \bibinfo{author}{Roy, A.},
  \bibinfo{author}{Patel, V.M.}, \bibinfo{year}{2022}.
\newblock \bibinfo{title}{Multimodal learning using optimal transport for
  sarcasm and humor detection}, in: \bibinfo{booktitle}{Proceedings of the
  IEEE/CVF Winter Conference on Applications of Computer Vision (WACV)}, pp.
  \bibinfo{pages}{3930--3940}.
%Type = Inproceedings
\bibitem[{Qiao et~al.(2021)Qiao, Yin, Wiechmann and Kerz}]{qiao21_interspeech}
\bibinfo{author}{Qiao, Y.}, \bibinfo{author}{Yin, X.},
  \bibinfo{author}{Wiechmann, D.}, \bibinfo{author}{Kerz, E.},
  \bibinfo{year}{2021}.
\newblock \bibinfo{title}{{Alzheimer’s Disease Detection from Spontaneous
  Speech Through Combining Linguistic Complexity and (Dis)Fluency Features with
  Pretrained Language Models}}, in: \bibinfo{booktitle}{Proc. Interspeech
  2021}, pp. \bibinfo{pages}{3805--3809}.
\newblock \DOIprefix\doi{10.21437/Interspeech.2021-1415}.
%Type = Article
\bibitem[{Radford et~al.(2022)Radford, Kim, Xu, Brockman, McLeavey and
  Sutskever}]{radford2022robust}
\bibinfo{author}{Radford, A.}, \bibinfo{author}{Kim, J.W.},
  \bibinfo{author}{Xu, T.}, \bibinfo{author}{Brockman, G.},
  \bibinfo{author}{McLeavey, C.}, \bibinfo{author}{Sutskever, I.},
  \bibinfo{year}{2022}.
\newblock \bibinfo{title}{Robust speech recognition via large-scale weak
  supervision}.
\newblock \bibinfo{journal}{arXiv preprint arXiv:2212.04356} .
%Type = Inproceedings
\bibitem[{Raghu et~al.(2019)Raghu, Blumer, Sayres, Obermeyer, Kleinberg,
  Mullainathan and Kleinberg}]{pmlr-v97-raghu19a}
\bibinfo{author}{Raghu, M.}, \bibinfo{author}{Blumer, K.},
  \bibinfo{author}{Sayres, R.}, \bibinfo{author}{Obermeyer, Z.},
  \bibinfo{author}{Kleinberg, B.}, \bibinfo{author}{Mullainathan, S.},
  \bibinfo{author}{Kleinberg, J.}, \bibinfo{year}{2019}.
\newblock \bibinfo{title}{Direct uncertainty prediction for medical second
  opinions}, in: \bibinfo{editor}{Chaudhuri, K.},
  \bibinfo{editor}{Salakhutdinov, R.} (Eds.), \bibinfo{booktitle}{Proceedings
  of the 36th International Conference on Machine Learning},
  \bibinfo{publisher}{PMLR}. pp. \bibinfo{pages}{5281--5290}.
\newblock \URLprefix \url{https://proceedings.mlr.press/v97/raghu19a.html}.
%Type = Article
\bibitem[{Reimers and Gurevych(2018)}]{reimers2018comparing}
\bibinfo{author}{Reimers, N.}, \bibinfo{author}{Gurevych, I.},
  \bibinfo{year}{2018}.
\newblock \bibinfo{title}{Why comparing single performance scores does not
  allow to draw conclusions about machine learning approaches}.
\newblock \bibinfo{journal}{arXiv preprint arXiv:1803.09578} .
%Type = Inproceedings
\bibitem[{Rohanian et~al.(2020)Rohanian, Hough and
  Purver}]{rohanian20_interspeech}
\bibinfo{author}{Rohanian, M.}, \bibinfo{author}{Hough, J.},
  \bibinfo{author}{Purver, M.}, \bibinfo{year}{2020}.
\newblock \bibinfo{title}{{Multi-Modal Fusion with Gating Using Audio, Lexical
  and Disfluency Features for Alzheimer’s Dementia Recognition from
  Spontaneous Speech}}, in: \bibinfo{booktitle}{Proc. Interspeech 2020}, pp.
  \bibinfo{pages}{2187--2191}.
\newblock \DOIprefix\doi{10.21437/Interspeech.2020-2721}.
%Type = Inproceedings
\bibitem[{Rohanian et~al.(2021)Rohanian, Hough and
  Purver}]{rohanian21_interspeech}
\bibinfo{author}{Rohanian, M.}, \bibinfo{author}{Hough, J.},
  \bibinfo{author}{Purver, M.}, \bibinfo{year}{2021}.
\newblock \bibinfo{title}{{Alzheimer’s Dementia Recognition Using Acoustic,
  Lexical, Disfluency and Speech Pause Features Robust to Noisy Inputs}}, in:
  \bibinfo{booktitle}{Proc. Interspeech 2021}, pp. \bibinfo{pages}{3820--3824}.
\newblock \DOIprefix\doi{10.21437/Interspeech.2021-1633}.
%Type = Inproceedings
\bibitem[{S{\'a}nchez~Villegas and
  Aletras(2021)}]{sanchez-villegas-aletras-2021-point}
\bibinfo{author}{S{\'a}nchez~Villegas, D.}, \bibinfo{author}{Aletras, N.},
  \bibinfo{year}{2021}.
\newblock \bibinfo{title}{Point-of-interest type prediction using text and
  images}, in: \bibinfo{booktitle}{Proceedings of the 2021 Conference on
  Empirical Methods in Natural Language Processing},
  \bibinfo{publisher}{Association for Computational Linguistics},
  \bibinfo{address}{Online and Punta Cana, Dominican Republic}. pp.
  \bibinfo{pages}{7785--7797}.
\newblock \URLprefix \url{https://aclanthology.org/2021.emnlp-main.614},
  \DOIprefix\doi{10.18653/v1/2021.emnlp-main.614}.
%Type = Inproceedings
\bibitem[{S{\'a}nchez~Villegas et~al.(2021)S{\'a}nchez~Villegas, Mokaram and
  Aletras}]{sanchez-villegas-etal-2021-analyzing}
\bibinfo{author}{S{\'a}nchez~Villegas, D.}, \bibinfo{author}{Mokaram, S.},
  \bibinfo{author}{Aletras, N.}, \bibinfo{year}{2021}.
\newblock \bibinfo{title}{Analyzing online political advertisements}, in:
  \bibinfo{booktitle}{Findings of the Association for Computational
  Linguistics: ACL-IJCNLP 2021}, \bibinfo{publisher}{Association for
  Computational Linguistics}, \bibinfo{address}{Online}. pp.
  \bibinfo{pages}{3669--3680}.
\newblock \URLprefix \url{https://aclanthology.org/2021.findings-acl.321},
  \DOIprefix\doi{10.18653/v1/2021.findings-acl.321}.
%Type = Inproceedings
\bibitem[{Sarawgi et~al.(2020)Sarawgi, Zulfikar, Soliman and
  Maes}]{sarawgi20_interspeech}
\bibinfo{author}{Sarawgi, U.}, \bibinfo{author}{Zulfikar, W.},
  \bibinfo{author}{Soliman, N.}, \bibinfo{author}{Maes, P.},
  \bibinfo{year}{2020}.
\newblock \bibinfo{title}{{Multimodal Inductive Transfer Learning for Detection
  of Alzheimer’s Dementia and its Severity}}, in: \bibinfo{booktitle}{Proc.
  Interspeech 2020}, pp. \bibinfo{pages}{2212--2216}.
\newblock \DOIprefix\doi{10.21437/Interspeech.2020-3137}.
%Type = Article
\bibitem[{Shah et~al.(2021)Shah, Sawalha, Tasnim, Qi, Stroulia and
  Greiner}]{10.3389/fcomp.2021.624659}
\bibinfo{author}{Shah, Z.}, \bibinfo{author}{Sawalha, J.},
  \bibinfo{author}{Tasnim, M.}, \bibinfo{author}{Qi, S.a.},
  \bibinfo{author}{Stroulia, E.}, \bibinfo{author}{Greiner, R.},
  \bibinfo{year}{2021}.
\newblock \bibinfo{title}{Learning language and acoustic models for identifying
  alzheimer’s dementia from speech}.
\newblock \bibinfo{journal}{Frontiers in Computer Science} \bibinfo{volume}{3},
  \bibinfo{pages}{4}.
\newblock \URLprefix
  \url{https://www.frontiersin.org/article/10.3389/fcomp.2021.624659},
  \DOIprefix\doi{10.3389/fcomp.2021.624659}.
%Type = Article
\bibitem[{Syed et~al.(2021a)Syed, Syed, Lech and Pirogova}]{9459113}
\bibinfo{author}{Syed, Z.S.}, \bibinfo{author}{Syed, M.S.S.},
  \bibinfo{author}{Lech, M.}, \bibinfo{author}{Pirogova, E.},
  \bibinfo{year}{2021}a.
\newblock \bibinfo{title}{Automated recognition of alzheimer’s dementia using
  bag-of-deep-features and model ensembling}.
\newblock \bibinfo{journal}{IEEE Access} \bibinfo{volume}{9},
  \bibinfo{pages}{88377--88390}.
\newblock \DOIprefix\doi{10.1109/ACCESS.2021.3090321}.
%Type = Inproceedings
\bibitem[{Syed et~al.(2021b)Syed, Syed, Lech and Pirogova}]{syed21_interspeech}
\bibinfo{author}{Syed, Z.S.}, \bibinfo{author}{Syed, M.S.S.},
  \bibinfo{author}{Lech, M.}, \bibinfo{author}{Pirogova, E.},
  \bibinfo{year}{2021}b.
\newblock \bibinfo{title}{{Tackling the ADRESSO Challenge 2021: The MUET-RMIT
  System for Alzheimer’s Dementia Recognition from Spontaneous Speech}}, in:
  \bibinfo{booktitle}{Proc. Interspeech 2021}, pp. \bibinfo{pages}{3815--3819}.
\newblock \DOIprefix\doi{10.21437/Interspeech.2021-1572}.
%Type = Inproceedings
\bibitem[{Szegedy et~al.(2016)Szegedy, Vanhoucke, Ioffe, Shlens and
  Wojna}]{7780677}
\bibinfo{author}{Szegedy, C.}, \bibinfo{author}{Vanhoucke, V.},
  \bibinfo{author}{Ioffe, S.}, \bibinfo{author}{Shlens, J.},
  \bibinfo{author}{Wojna, Z.}, \bibinfo{year}{2016}.
\newblock \bibinfo{title}{Rethinking the inception architecture for computer
  vision}, in: \bibinfo{booktitle}{2016 IEEE Conference on Computer Vision and
  Pattern Recognition (CVPR)}, pp. \bibinfo{pages}{2818--2826}.
\newblock \DOIprefix\doi{10.1109/CVPR.2016.308}.
%Type = Inproceedings
\bibitem[{Tan and Le(2019)}]{tan2019efficientnet}
\bibinfo{author}{Tan, M.}, \bibinfo{author}{Le, Q.}, \bibinfo{year}{2019}.
\newblock \bibinfo{title}{Efficientnet: Rethinking model scaling for
  convolutional neural networks}, in: \bibinfo{booktitle}{International
  Conference on Machine Learning}, \bibinfo{organization}{PMLR}. pp.
  \bibinfo{pages}{6105--6114}.
%Type = Inproceedings
\bibitem[{Touvron et~al.(2021)Touvron, Cord, Douze, Massa, Sablayrolles and
  Jegou}]{pmlr-v139-touvron21a}
\bibinfo{author}{Touvron, H.}, \bibinfo{author}{Cord, M.},
  \bibinfo{author}{Douze, M.}, \bibinfo{author}{Massa, F.},
  \bibinfo{author}{Sablayrolles, A.}, \bibinfo{author}{Jegou, H.},
  \bibinfo{year}{2021}.
\newblock \bibinfo{title}{Training data-efficient image transformers \&amp;
  distillation through attention}, in: \bibinfo{editor}{Meila, M.},
  \bibinfo{editor}{Zhang, T.} (Eds.), \bibinfo{booktitle}{Proceedings of the
  38th International Conference on Machine Learning},
  \bibinfo{publisher}{PMLR}. pp. \bibinfo{pages}{10347--10357}.
\newblock \URLprefix \url{https://proceedings.mlr.press/v139/touvron21a.html}.
%Type = Inproceedings
\bibitem[{Tsai et~al.(2019)Tsai, Bai, Liang, Kolter, Morency and
  Salakhutdinov}]{tsai-etal-2019-multimodal}
\bibinfo{author}{Tsai, Y.H.H.}, \bibinfo{author}{Bai, S.},
  \bibinfo{author}{Liang, P.P.}, \bibinfo{author}{Kolter, J.Z.},
  \bibinfo{author}{Morency, L.P.}, \bibinfo{author}{Salakhutdinov, R.},
  \bibinfo{year}{2019}.
\newblock \bibinfo{title}{Multimodal transformer for unaligned multimodal
  language sequences}, in: \bibinfo{booktitle}{Proceedings of the 57th Annual
  Meeting of the Association for Computational Linguistics},
  \bibinfo{publisher}{Association for Computational Linguistics},
  \bibinfo{address}{Florence, Italy}. pp. \bibinfo{pages}{6558--6569}.
\newblock \URLprefix \url{https://aclanthology.org/P19-1656},
  \DOIprefix\doi{10.18653/v1/P19-1656}.
%Type = Article
\bibitem[{Tu et~al.(2017)Tu, Liu, Lu, Liu and Li}]{tu-etal-2017-context}
\bibinfo{author}{Tu, Z.}, \bibinfo{author}{Liu, Y.}, \bibinfo{author}{Lu, Z.},
  \bibinfo{author}{Liu, X.}, \bibinfo{author}{Li, H.}, \bibinfo{year}{2017}.
\newblock \bibinfo{title}{Context gates for neural machine translation}.
\newblock \bibinfo{journal}{Transactions of the Association for Computational
  Linguistics} \bibinfo{volume}{5}, \bibinfo{pages}{87--99}.
\newblock \URLprefix \url{https://aclanthology.org/Q17-1007},
  \DOIprefix\doi{10.1162/tacl_a_00048}.
%Type = Article
\bibitem[{Ulmer et~al.(2022)Ulmer, Hardmeier and Frellsen}]{ulmer2022deep}
\bibinfo{author}{Ulmer, D.}, \bibinfo{author}{Hardmeier, C.},
  \bibinfo{author}{Frellsen, J.}, \bibinfo{year}{2022}.
\newblock \bibinfo{title}{deep-significance-easy and meaningful statistical
  significance testing in the age of neural networks}.
\newblock \bibinfo{journal}{arXiv preprint arXiv:2204.06815} .
%Type = Inproceedings
\bibitem[{Vaswani et~al.(2017)Vaswani, Shazeer, Parmar, Uszkoreit, Jones,
  Gomez, Kaiser and Polosukhin}]{10.5555/3295222.3295349}
\bibinfo{author}{Vaswani, A.}, \bibinfo{author}{Shazeer, N.},
  \bibinfo{author}{Parmar, N.}, \bibinfo{author}{Uszkoreit, J.},
  \bibinfo{author}{Jones, L.}, \bibinfo{author}{Gomez, A.N.},
  \bibinfo{author}{Kaiser, L.}, \bibinfo{author}{Polosukhin, I.},
  \bibinfo{year}{2017}.
\newblock \bibinfo{title}{Attention is all you need}, in:
  \bibinfo{booktitle}{Proceedings of the 31st International Conference on
  Neural Information Processing Systems}, \bibinfo{publisher}{Curran Associates
  Inc.}, \bibinfo{address}{Red Hook, NY, USA}. p. \bibinfo{pages}{6000–6010}.
%Type = Article
\bibitem[{Villani(2008)}]{villani2008optimal}
\bibinfo{author}{Villani, C.}, \bibinfo{year}{2008}.
\newblock \bibinfo{title}{Optimal transport, old and new. notes for the 2005
  saint-flour summer school}.
\newblock \bibinfo{journal}{Grundlehren der mathematischen Wissenschaften
  [Fundamental Principles of Mathematical Sciences]. Springer} .
%Type = Inproceedings
\bibitem[{Voita et~al.(2018)Voita, Serdyukov, Sennrich and
  Titov}]{voita-etal-2018-context}
\bibinfo{author}{Voita, E.}, \bibinfo{author}{Serdyukov, P.},
  \bibinfo{author}{Sennrich, R.}, \bibinfo{author}{Titov, I.},
  \bibinfo{year}{2018}.
\newblock \bibinfo{title}{Context-aware neural machine translation learns
  anaphora resolution}, in: \bibinfo{booktitle}{Proceedings of the 56th Annual
  Meeting of the Association for Computational Linguistics (Volume 1: Long
  Papers)}, \bibinfo{publisher}{Association for Computational Linguistics},
  \bibinfo{address}{Melbourne, Australia}. pp. \bibinfo{pages}{1264--1274}.
\newblock \URLprefix \url{https://aclanthology.org/P18-1117},
  \DOIprefix\doi{10.18653/v1/P18-1117}.
%Type = Inproceedings
\bibitem[{Wang et~al.(2017)Wang, Tu, Way and
  Liu}]{wang-etal-2017-exploiting-cross}
\bibinfo{author}{Wang, L.}, \bibinfo{author}{Tu, Z.}, \bibinfo{author}{Way,
  A.}, \bibinfo{author}{Liu, Q.}, \bibinfo{year}{2017}.
\newblock \bibinfo{title}{Exploiting cross-sentence context for neural machine
  translation}, in: \bibinfo{booktitle}{Proceedings of the 2017 Conference on
  Empirical Methods in Natural Language Processing},
  \bibinfo{publisher}{Association for Computational Linguistics},
  \bibinfo{address}{Copenhagen, Denmark}. pp. \bibinfo{pages}{2826--2831}.
\newblock \URLprefix \url{https://aclanthology.org/D17-1301},
  \DOIprefix\doi{10.18653/v1/D17-1301}.
%Type = Inproceedings
\bibitem[{Wang et~al.(2021)Wang, Cao, Hao, Shao and
  Subbalakshmi}]{wang21ca_interspeech}
\bibinfo{author}{Wang, N.}, \bibinfo{author}{Cao, Y.}, \bibinfo{author}{Hao,
  S.}, \bibinfo{author}{Shao, Z.}, \bibinfo{author}{Subbalakshmi, K.},
  \bibinfo{year}{2021}.
\newblock \bibinfo{title}{{Modular Multi-Modal Attention Network for
  Alzheimer’s Disease Detection Using Patient Audio and Language Data}}, in:
  \bibinfo{booktitle}{Proc. Interspeech 2021}, pp. \bibinfo{pages}{3835--3839}.
\newblock \DOIprefix\doi{10.21437/Interspeech.2021-2024}.
%Type = Inproceedings
\bibitem[{Wolf et~al.(2020)Wolf, Debut, Sanh, Chaumond, Delangue, Moi, Cistac,
  Rault, Louf, Funtowicz, Davison, Shleifer, von Platen, Ma, Jernite, Plu, Xu,
  Scao, Gugger, Drame, Lhoest and Rush}]{wolf-etal-2020-transformers}
\bibinfo{author}{Wolf, T.}, \bibinfo{author}{Debut, L.}, \bibinfo{author}{Sanh,
  V.}, \bibinfo{author}{Chaumond, J.}, \bibinfo{author}{Delangue, C.},
  \bibinfo{author}{Moi, A.}, \bibinfo{author}{Cistac, P.},
  \bibinfo{author}{Rault, T.}, \bibinfo{author}{Louf, R.},
  \bibinfo{author}{Funtowicz, M.}, \bibinfo{author}{Davison, J.},
  \bibinfo{author}{Shleifer, S.}, \bibinfo{author}{von Platen, P.},
  \bibinfo{author}{Ma, C.}, \bibinfo{author}{Jernite, Y.},
  \bibinfo{author}{Plu, J.}, \bibinfo{author}{Xu, C.}, \bibinfo{author}{Scao,
  T.L.}, \bibinfo{author}{Gugger, S.}, \bibinfo{author}{Drame, M.},
  \bibinfo{author}{Lhoest, Q.}, \bibinfo{author}{Rush, A.M.},
  \bibinfo{year}{2020}.
\newblock \bibinfo{title}{Transformers: State-of-the-art natural language
  processing}, in: \bibinfo{booktitle}{Proceedings of the 2020 Conference on
  Empirical Methods in Natural Language Processing: System Demonstrations},
  \bibinfo{publisher}{Association for Computational Linguistics},
  \bibinfo{address}{Online}. pp. \bibinfo{pages}{38--45}.
\newblock \URLprefix \url{https://www.aclweb.org/anthology/2020.emnlp-demos.6}.
%Type = Misc
\bibitem[{{World Health Organization}({2021})}]{alzheimer_who}
\bibinfo{author}{{World Health Organization}}, \bibinfo{year}{{2021}}.
\newblock \bibinfo{title}{{\textit{Dementia}}}.
\newblock \bibinfo{howpublished}{Available online at:
  \url{https://www.who.int/news-room/fact-sheets/detail/dementia}}.
%Type = Article
\bibitem[{Yang et~al.(2019)Yang, Li, Wong, Chao, Wang and
  Tu}]{Yang_Li_Wong_Chao_Wang_Tu_2019}
\bibinfo{author}{Yang, B.}, \bibinfo{author}{Li, J.}, \bibinfo{author}{Wong,
  D.F.}, \bibinfo{author}{Chao, L.S.}, \bibinfo{author}{Wang, X.},
  \bibinfo{author}{Tu, Z.}, \bibinfo{year}{2019}.
\newblock \bibinfo{title}{Context-aware self-attention networks}.
\newblock \bibinfo{journal}{Proceedings of the AAAI Conference on Artificial
  Intelligence} \bibinfo{volume}{33}, \bibinfo{pages}{387--394}.
\newblock \URLprefix
  \url{https://ojs.aaai.org/index.php/AAAI/article/view/3809},
  \DOIprefix\doi{10.1609/aaai.v33i01.3301387}.
%Type = Inproceedings
\bibitem[{Yang et~al.(2022)Yang, Wei, Li, Li and
  Shinozaki}]{yang22k_interspeech}
\bibinfo{author}{Yang, L.}, \bibinfo{author}{Wei, W.}, \bibinfo{author}{Li,
  S.}, \bibinfo{author}{Li, J.}, \bibinfo{author}{Shinozaki, T.},
  \bibinfo{year}{2022}.
\newblock \bibinfo{title}{{Augmented Adversarial Self-Supervised Learning for
  Early-Stage Alzheimer's Speech Detection}}, in: \bibinfo{booktitle}{Proc.
  Interspeech 2022}, pp. \bibinfo{pages}{541--545}.
\newblock \DOIprefix\doi{10.21437/Interspeech.2022-943}.
%Type = Article
\bibitem[{Yu et~al.(2019a)Yu, Cui, Yu, Tao and Tian}]{yu2019multimodal}
\bibinfo{author}{Yu, Z.}, \bibinfo{author}{Cui, Y.}, \bibinfo{author}{Yu, J.},
  \bibinfo{author}{Tao, D.}, \bibinfo{author}{Tian, Q.}, \bibinfo{year}{2019}a.
\newblock \bibinfo{title}{Multimodal unified attention networks for
  vision-and-language interactions}.
\newblock \bibinfo{journal}{arXiv preprint arXiv:1908.04107} .
%Type = Inproceedings
\bibitem[{Yu et~al.(2019b)Yu, Yu, Cui, Tao and Tian}]{Yu_2019_CVPR}
\bibinfo{author}{Yu, Z.}, \bibinfo{author}{Yu, J.}, \bibinfo{author}{Cui, Y.},
  \bibinfo{author}{Tao, D.}, \bibinfo{author}{Tian, Q.}, \bibinfo{year}{2019}b.
\newblock \bibinfo{title}{Deep modular co-attention networks for visual
  question answering}, in: \bibinfo{booktitle}{Proceedings of the IEEE/CVF
  Conference on Computer Vision and Pattern Recognition (CVPR)}.
%Type = Inproceedings
\bibitem[{Zadeh et~al.(2017)Zadeh, Chen, Poria, Cambria and
  Morency}]{zadeh-etal-2017-tensor}
\bibinfo{author}{Zadeh, A.}, \bibinfo{author}{Chen, M.},
  \bibinfo{author}{Poria, S.}, \bibinfo{author}{Cambria, E.},
  \bibinfo{author}{Morency, L.P.}, \bibinfo{year}{2017}.
\newblock \bibinfo{title}{Tensor fusion network for multimodal sentiment
  analysis}, in: \bibinfo{booktitle}{Proceedings of the 2017 Conference on
  Empirical Methods in Natural Language Processing},
  \bibinfo{publisher}{Association for Computational Linguistics},
  \bibinfo{address}{Copenhagen, Denmark}. pp. \bibinfo{pages}{1103--1114}.
\newblock \URLprefix \url{https://aclanthology.org/D17-1115},
  \DOIprefix\doi{10.18653/v1/D17-1115}.
%Type = Article
\bibitem[{Zhang et~al.(2017)Zhang, Xiong, Su and Duan}]{8031316}
\bibinfo{author}{Zhang, B.}, \bibinfo{author}{Xiong, D.}, \bibinfo{author}{Su,
  J.}, \bibinfo{author}{Duan, H.}, \bibinfo{year}{2017}.
\newblock \bibinfo{title}{A context-aware recurrent encoder for neural machine
  translation}.
\newblock \bibinfo{journal}{IEEE/ACM Transactions on Audio, Speech, and
  Language Processing} \bibinfo{volume}{25}, \bibinfo{pages}{2424--2432}.
\newblock \DOIprefix\doi{10.1109/TASLP.2017.2751420}.
%Type = Article
\bibitem[{Zhu et~al.(2021a)Zhu, Liang, Batsis and
  Roth}]{10.3389/fcomp.2021.624683}
\bibinfo{author}{Zhu, Y.}, \bibinfo{author}{Liang, X.},
  \bibinfo{author}{Batsis, J.A.}, \bibinfo{author}{Roth, R.M.},
  \bibinfo{year}{2021}a.
\newblock \bibinfo{title}{Exploring deep transfer learning techniques for
  alzheimer's dementia detection}.
\newblock \bibinfo{journal}{Frontiers in Computer Science} \bibinfo{volume}{3},
  \bibinfo{pages}{22}.
\newblock \URLprefix
  \url{https://www.frontiersin.org/article/10.3389/fcomp.2021.624683},
  \DOIprefix\doi{10.3389/fcomp.2021.624683}.
%Type = Inproceedings
\bibitem[{Zhu et~al.(2021b)Zhu, Obyat, Liang, Batsis and
  Roth}]{zhu21e_interspeech}
\bibinfo{author}{Zhu, Y.}, \bibinfo{author}{Obyat, A.}, \bibinfo{author}{Liang,
  X.}, \bibinfo{author}{Batsis, J.A.}, \bibinfo{author}{Roth, R.M.},
  \bibinfo{year}{2021}b.
\newblock \bibinfo{title}{{WavBERT: Exploiting Semantic and Non-Semantic Speech
  Using Wav2vec and BERT for Dementia Detection}}, in:
  \bibinfo{booktitle}{Proc. Interspeech 2021}, pp. \bibinfo{pages}{3790--3794}.
\newblock \DOIprefix\doi{10.21437/Interspeech.2021-332}.

\end{thebibliography}



\end{document}